\documentclass[11pt]{article}

\usepackage{acl}

\usepackage{times}
\usepackage{latexsym}

\usepackage[T1]{fontenc}

\usepackage[utf8]{inputenc}

\usepackage{microtype}

\usepackage{inconsolata}

\usepackage{graphicx}

\usepackage{amsmath}
\usepackage{amssymb}
\usepackage{booktabs}
\usepackage{multirow}
\usepackage{colortbl}
\usepackage{wrapfig}
\usepackage{algorithm}
\usepackage{algpseudocodex}
\usepackage{pifont}
\usepackage[breakable]{tcolorbox}
\usepackage{xcolor}
\definecolor{mypink}{RGB}{255,223,234}  
\definecolor{myframepink}{RGB}{245,104,138}  
\definecolor{citepink}{RGB}{224,80,128}  
\usepackage{enumitem}
\usepackage{tabularx}
\usepackage{array}

\usepackage{hyperref}       

\hypersetup{
    colorlinks=true,       
    allcolors=myframepink  
}

\title{iTAG: Inverse Design for Natural Text Generation with Accurate Causal Graph Annotations}

\author{
  Wenshuo Wang$^{1}$ \quad Boyu Cao$^{1}$ \quad Nan Zhuang$^{2}$ \quad Wei Li$^{3,\dagger}$\\
  $^{1}$School of Future Technology, South China University of Technology, Guangzhou, China\\
  $^{2}$China Mobile GBA Innovation Institute, Guangzhou, China\\
  $^{3}$Institute of Computing Technology, Chinese Academy of Sciences, Beijing, China\\
  \small{$^{\dagger}$Corresponding author: \href{mailto:liwei@ict.ac.cn}{liwei@ict.ac.cn}}
}

\begin{document}
\maketitle

\begin{abstract}
A fundamental obstacle to causal discovery from text is the lack of causally annotated text data for use as ground truth, due to high annotation costs. This motivates an important task of generating text with causal graph annotations. Early template-based generation methods sacrifice text naturalness in exchange for high causal graph annotation accuracy. Recent Large Language Model (LLM)-dependent methods directly generate natural text from target graphs through LLMs, but do not guarantee causal graph annotation accuracy. Therefore, we propose iTAG, which performs real-world concept assignment to nodes before converting causal graphs into text in existing LLM-dependent methods. iTAG frames this process as an inverse problem with the causal graph as the target, iteratively examining and refining concept selection through Chain-of-Thought (CoT) reasoning so that the induced relations between concepts are as consistent as possible with the target causal relationships described by the causal graph. iTAG demonstrates both extremely high annotation accuracy and naturalness across extensive tests, and the results of testing text-based causal discovery algorithms with the generated data show high statistical correlation with real-world data. This suggests that iTAG-generated data can serve as a practical surrogate for scalable benchmarking of text-based causal discovery algorithms.
\end{abstract}

\section{Introduction}

Due to the extremely high cost of causally annotating real-world data, there is often a lack of annotated real-world data for validating the effectiveness of causal discovery algorithms \citep{faller2024self, hiremath2024losam}. Therefore, synthetic data with controllable parameters is widely used in causal discovery research as ideal ground truth for reproducible algorithm validation \citep{brouillard2024landscape}. However, for generating causally annotated text, causal relationships are implicit in semantics and knowledge \citep{ding2025multi}. This makes them difficult to explicitly control and annotate, thus generating causally annotated text is more challenging. (Throughout this paper, by \emph{text-based causal discovery} we mean evaluating whether a downstream method can recover a target adjacency-level causal structure from text with known ground truth, rather than discovering previously unknown scientific causal relations beyond an LLM's knowledge.) This difficulty has gradually become one of the fundamental obstacles to the development of causal discovery from text \citep{wang2025causalenhance, gujarathi2022study}.

Earlier methods for generating causally annotated text adopt template-based approaches (e.g., "[A] results in [B]"), which can ensure the accuracy of causal descriptions yet generate text that is unnatural \citep{shrestha2022automatically}. Recent generation methods directly convert predefined causal graphs into natural text through LLMs, but do not verify whether the concepts described in the generated text conform to the target causal relationships \citep{phatak2024narrating, Gandee, gandee2025faithful}. In summary, text generated by current methods is either unnatural or has inaccurate causal graph annotations, and cannot serve as a credible substitute for real human-annotated data.

We propose \textbf{iTAG}: \textbf{i}nverse design for \textbf{T}ext gener\textbf{A}tion with causal \textbf{G}raph. Building upon existing LLM-dependent methods, iTAG first performs real-world concept assignment to nodes before converting causal graphs into text. We frame this process as an inverse problem of designing real-world concepts with the causal graph as the target. LLMs serve as solvers that iteratively examine and refine concept selection through CoT until the relationships between concepts are judged to be consistent with the target causal relationships.

We extensively validated the extremely high annotation accuracy of iTAG across text data of multiple topics. We also verified the naturalness of generated text through both human evaluation and trained model discrimination. Furthermore, evaluations of representative text-based causal discovery algorithms on iTAG-generated corpora are highly predictive of their evaluations on matched real-world corpora. This suggests that iTAG-generated data can serve as a practical surrogate for scalable benchmarking when human causal annotations are scarce (while not replacing real-world evaluation).

In summary, our main contributions include:
\begin{itemize}
  \item \textbf{A Novel Generation Framework:} We propose iTAG, which introduces an additional concept assignment step compared to existing LLM-dependent pipelines, reformulating text generation as a concept design problem constrained by causal graphs.
  \item \textbf{Dual Excellence in Accuracy and Naturalness:} We simultaneously achieve extremely high causal graph annotation accuracy and text naturalness, breaking the trade-off dilemma of existing methods.
  \item \textbf{Validated Surrogate for Real-World Evaluation:} In application, algorithm rankings and relative performance measured on iTAG-generated corpora closely track those on real corpora across multiple causal discovery paradigms. This enables cheaper and more systematic benchmarking without large-scale manual causal annotation.
\end{itemize}

\section{Related Work}

\begin{figure*}
  \centering
  \includegraphics[width=\linewidth]{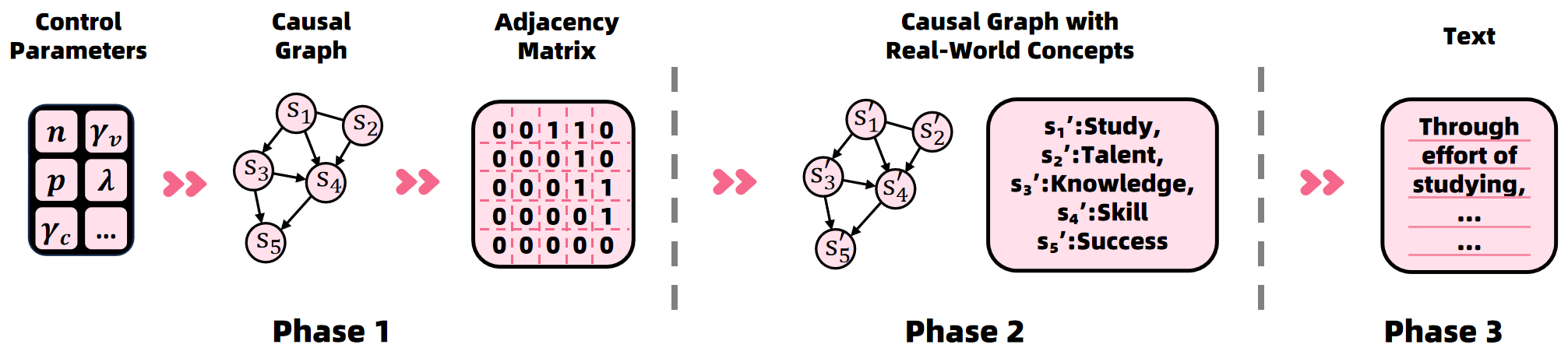}
  \caption{An example of the three-phase workflow of \textbf{iTAG}: \textbf{i}nverse design for \textbf{T}ext gener\textbf{A}tion with causal \textbf{G}raph.}
  \label{fig:methodsim}
\end{figure*}

\subsection{Text generation with causal graph annotations}\label{sec:rw_generation}

Text generation with causal graph annotations is a task that transforms causal graphs into natural language text while preserving all causal relationships. Formally, given a causal graph $G = (V, E)$, where $V = \{v_1, v_2, ..., v_n\}$ represents concept nodes and $E \subseteq V \times V$ represents directed causal relationships such that $(v_i, v_j) \in E$ indicates $v_i$ causally influences $v_j$, the objective is to generate text $T$ comprising a set of sentences $S = \{s_1, s_2, ..., s_m\}$ that linguistically encode all relationships in $E$ without introducing spurious connections not present in $G$. The task involves both generating the text $T$ and establishing an annotation function $A$ that maps $T$ to a reconstructed causal graph $G' = A(T)$, where ideally $G'$ is isomorphic to $G$.

Earlier methods predominantly use template or rule-driven approaches to transcribe target causal relationships one by one into sentences and concatenate them into paragraphs. This achieves high coverage but with fixed discourse structure and rigid expression \citep{shrestha2022automatically}. Recent work directly chunks or linearizes predefined causal graphs and inputs them into LLMs to generate text, without additionally setting up a solving step for assigning concepts to nodes. Moreover, they typically do not introduce consistency verification based on whole-graph binary constraints during generation, only performing post-hoc evaluation or using graph algorithms to assist in organizing paragraphs. Therefore, the authors also point out that information omission or hallucination may occur \citep{phatak2024narrating, Gandee, gandee2025faithful}.

Accordingly, template-based methods provide better guarantees in annotation coverage and controllability, but sacrifice text naturalness and coherence \citep{shrestha2022automatically}. In contrast, direct LLM conversion achieves better naturalness, but commonly suffers from omission and hallucination risks due to the lack of global hard constraints during generation. This leads to unstable accuracy of causal annotations recovered from graphs \citep{phatak2024narrating, Gandee, gandee2025faithful}. In summary, text generated by current methods is either unnatural or has inaccurate causal graph annotations, and cannot serve as a credible substitute for real human-annotated data.

\subsection{Inverse design and CoT}

To our knowledge, iTAG is the first to apply an inverse-design-based iterative refinement loop to LLM-based concept assignment under whole-graph causal constraints. The introduction of these key concepts is as follows: \textbf{Inverse problems} refer to inferring unknown parameters from indirect and noisy observations given a known forward model \citep{stuart2010inverse}. \textbf{Inverse design} parameterizes design spaces and optimizes performance objectives by iteratively simulating candidates and updating parameters to minimize target-output gaps \citep{pfaff2020learning, kim2019deep}. \textbf{CoT} prompting is a prompt engineering technique that enables LLMs to perform complex reasoning by explicitly generating intermediate steps before reaching a final answer \citep{wei2023chainofthoughtpromptingelicitsreasoning,wang2022self}. Current research demonstrates that CoT prompting achieves substantial performance improvements, with accuracy gains across various reasoning benchmarks \citep{yao2023tree,zhou2023leasttomostpromptingenablescomplex}.

Existing LLM-dependent methods overly rely on LLMs' intrinsic reasoning capabilities, while well-designed CoT can significantly enhance reasoning abilities to resolve potential causal structure errors \citep{wei2023chainofthoughtpromptingelicitsreasoning}. Therefore, iTAG formulates concept assignment as an inverse problem and introduces inverse design into the assignment process through CoT. This iterative error-correction inverse design approach aims to improve causal structure accuracy by guiding LLM reasoning with explicit graph-level constraints, thereby addressing the dual challenges presented in Section~\ref{sec:rw_generation}.

\section{Method}\label{sec:method}

In this section, we first outline the three-phase workflow of iTAG in Section~\ref{sec:flow}, then detail its phases in Sections~\ref{sec:graph}, ~\ref{sec:inverse}, and ~\ref{sec:text}, respectively.

\subsection{Overview of the three-phase workflow of iTAG}\label{sec:flow}

iTAG generates text with causal graphs through a three-phase pipeline as shown in Figure~\ref{fig:methodsim}. In \textbf{phase 1}, \textbf{control parameters} such as variable quantity ($n$), expected graph density ($p$), collider ratio ($\gamma_v$), mediator chains count ($\lambda$), and confounder ratio ($\gamma_c$) are transformed into a structured \textbf{causal graph} (nodes $s_1$ through $s_5$ in the example) and subsequently converted into an \textbf{adjacency matrix}. This matrix precisely encodes all causal relationships, with entries of 1 indicating direct causal influences (such as $s_1 \rightarrow s_3$ in the example) while 0s represent the absence of such relationships. In \textbf{phase 2}, abstract variables in the causal graph undergo substitution with \textbf{real-world concepts} ($s_1'$:"Study", $s_2'$:"Talent", $s_3'$:"Knowledge", $s_4'$:"Skill", $s_5'$:"Success") while aiming to preserve the causal structure defined in the adjacency matrix. In \textbf{phase 3}, these real-world concepts and their causal structure are transformed into coherent natural language \textbf{text} that implicitly embeds the defined causal relationships, such as the text generated in the example:

\begin{tcolorbox}[colback=mypink, colframe=myframepink]
Through the effort of \textbf{studying}, individuals acquire \textbf{knowledge} while developing \textbf{skills} enhanced by their natural \textbf{talents}. As \textbf{knowledge} accumulates, it supports the development and refinement of \textbf{skills}, and those who possess both \textbf{knowledge} and refined \textbf{skills} typically achieve \textbf{success}.
\end{tcolorbox}

\subsection{Phase 1: parameterized causal graph construction}\label{sec:graph}

\begin{algorithm*}[t]
\caption{inverse-design-based concept substitution}
\label{alg:concept_substitution}
\begin{algorithmic}[1]
\Require Adjacency matrix $A$, max iterations $K_{\max}$ \; (fixed verifier settings; Appendix~\ref{sec:appendix_itag_verifier_protocol})
\Ensure Concept assignment $C$ and status $s \in \{\textsc{Success}, \textsc{Fail}\}$
\State $R \gets \textsc{AnalyzeCausalStructure}(A)$; \;\; $C \gets \textsc{InitialConceptAssignment}()$
\State $C^\star \gets C$; \;\; $L^\star \gets +\infty$
\For{$t = 1$ to $K_{\max}$}
  \State $V \gets \textsc{CounterfactualVerification}(C, R)$
  \State $F \gets \textsc{FallacyAnalysis}(V)$; \;\; $L \gets \textsc{QuantifyMismatch}(V, A)$
  \If{$L < L^\star$} $L^\star \gets L$; $C^\star \gets C$ \EndIf
  \If{$F = \emptyset$} \Return $(C, \textsc{Success})$ \EndIf
  \State $C \gets \textsc{RefineConceptAssignment}(C, F)$
\EndFor
\State \Return $(C^\star, \textsc{Fail})$
\end{algorithmic}
\end{algorithm*}

Phase~1 transforms control parameters into structured causal graphs and adjacency matrices. The \textbf{input} includes the variable quantity $n$, expected density $p$, degree limits ($max\_parents$, $max\_children$), and motif controls (confounder ratio $\gamma_c$, collider ratio $\gamma_v$, mediator-chain count $\lambda$), providing explicit control over structural complexity; the \textbf{output} is a DAG and its adjacency matrix $A$, where $a_{ij}=1$ denotes a direct edge $i\!\rightarrow\!j$ and $a_{ij}=0$ denotes its absence.

To execute this transformation, we implement an enhanced Erd\H{o}s--R\'enyi DAG generator \citep{erdds1959random, erd6s1960evolution}. We first sample edges to match the target density $p$ while enforcing $max\_parents$ and $max\_children$, and then inject $\gamma_c \!\times\! n$ confounders, $\gamma_v \!\times\! n$ colliders, and $\lambda$ mediator chains via dedicated subroutines.

\subsection{Phase 2: inverse-design-based concept substitution}\label{sec:inverse}

Phase~2 transforms abstract causal-graph nodes into domain-grounded real-world concepts while aiming to preserve the causal structure specified by the adjacency matrix. Here, a \emph{concept} means a node's real-world instantiation (e.g., an entity, variable, or noun phrase), rather than an abstract node ID. The \textbf{input} is the adjacency matrix $A$ from Phase~1, and the \textbf{output} is a concept assignment $C=(c_1,\dots,c_n)$ (subject to a non-overlap / granularity / intervention-meaningful schema; Appendix~\ref{sec:appendix_itag_schema}) that seeks to realize all required direct edges ($a_{ij}=1$) while suppressing spurious \emph{direct} relations on non-edges ($a_{ij}=0$). Throughout this phase, causal correctness follows four task-level conventions:
\begin{itemize}[leftmargin=1.2em, itemsep=1pt, topsep=2pt, parsep=0pt]
    \item each graph node is instantiated as one explicit concept in the text;
    \item each edge denotes a \emph{direct}, adjacency-level causal relation between two concepts;
    \item each non-edge means the text should not assert a \emph{direct} causal relation between the corresponding concepts; and
    \item the target graph contains all explicit concepts to be annotated in the generated text.
\end{itemize}
Since non-edges are hard to falsify from short narratives, we treat $a_{ij}=0$ as \emph{soft} evidence via graded penalties in $\widehat{\mathcal{L}}$ (Appendix~\ref{sec:appendix_itag_mismatch}); to reduce circularity, the verifier backbone is disjoint from the proposer/refiner by default (Appendix~\ref{sec:appendix_itag_llm_roles}).

Let $R\subseteq\{(i,j):i\neq j\}$ denote the set of ordered node pairs to be verified, and let $V=\{s_{ij}\}_{(i,j)\in R}$ denote the verifier outputs, where $s_{ij}\in[0,1]$ is the self-consistency vote proportion that $c_i$ is a \emph{direct} cause of $c_j$. In our implementation, $R=E^{+}\cup E^{-}$ with $E^{+}=\{(i,j):i\neq j,\ a_{ij}=1\}$ and $E^{-}=\{(i,j):i\neq j,\ a_{ij}=0\}$.
Given a candidate assignment $C$, \textsc{CounterfactualVerification} returns these verifier outputs; instantiating $\ell^{\mathrm{miss}}_{ij}=1-s_{ij}$ on $E^{+}$ and $\ell^{\mathrm{spur}}_{ij}=s_{ij}$ on $E^{-}$ yields the normalized diagnostic mismatch
\[
\begin{aligned}
\mathcal{L}(C; A)
= \sum_{i \neq j} \mathbf{1}[a_{ij} = 1] \cdot \ell^{\mathrm{miss}}_{ij}(C)
\\
\quad + \alpha \sum_{i \neq j} \mathbf{1}[a_{ij} = 0] \cdot \ell^{\mathrm{spur}}_{ij}(C),
\end{aligned}
\]
whose two terms correspond to missed-required vs.\ spurious-on-non-edge errors; importantly, the non-edge term penalizes only predicted \emph{direct} causality and does not assume perfect discrimination from indirect/confounded relations. For refinement, \textsc{FallacyAnalysis} additionally thresholds $s_{ij}$ to produce a violation set over $E^{+}\cup E^{-}$.

To execute this transformation, we treat concept assignment as an inverse-design problem, with the target graph $A$ as the desired output and the concept assignment $C$ as the design variables, and solve it with a CoT-guided propose--verify--refine loop (Algorithm~\ref{alg:concept_substitution}; prompt in Appendix~\ref{sec:prompt}): \textsc{AnalyzeCausalStructure} forms $E^{+}\cup E^{-}$, \textsc{InitialConceptAssignment} proposes $C^{(0)}$, \textsc{CounterfactualVerification} computes vote proportions, \textsc{QuantifyMismatch} scores $\widehat{\mathcal{L}}$, and \textsc{RefineConceptAssignment} revises $C$ based on the violation set while tracking the best-so-far $C^\star$ (minimum $\widehat{\mathcal{L}}$ over visited iterations). The loop returns \textsc{Success} when the violation set is empty; otherwise it runs for at most $K_{\max}$ iterations and returns $C^\star$ (\textsc{Fail}), \emph{without assuming any formal optimality guarantees}. Under the default fixed setting ($m=5$, $\tau=0.6$, $K_{\max}=10$, $\alpha=1$), the loop terminates in a median of 1.63 iterations with 99.1\% success rate and median verifier usage of 4.3k tokens per sample; for compactness and reproducibility, the appendix reports the full protocol/details and distributions (Appendix~\ref{sec:appendix_itag_verifier_protocol}--\ref{sec:appendix_itag_termination}) and a residual error decomposition / failure-mode analysis (Appendix~\ref{sec:appendix_error_decomp}).

\subsection{Phase 3: causal structure-preserving textual transformation}\label{sec:text}

Phase~3 transforms causal graphs with real-world concepts into natural language text while aiming to preserve the underlying causal structure. The \textbf{input} consists of the adjacency matrix from Phase~1 and the real-world concepts from Phase~2; the \textbf{output} is a coherent paragraph that aims to verbalize each required edge ($a_{ij}=1$) at least once and to avoid asserting \emph{direct} causality for non-edges ($a_{ij}=0$), although occasional omissions or ambiguous phrasings may still occur.

Given $(A,C)$, we enumerate parent--child pairs ($a_{ij}=1$) and prompt the LLM to weave them into fluent text while (i) forbidding any additional concepts beyond $C$ and (ii) discouraging direct causal assertions on $a_{ij}=0$ pairs; the concept-control prompts are in Appendix~\ref{sec:prompt}. Unlike Phase~2, Phase~3 uses a single-shot transformation rather than an additional inverse-design loop: once concepts are clear and non-overlapping, LLMs make few structural errors under these constraints, and a controlled ablation shows that adding a generation-time inverse-design loop yields only marginal gains at substantially higher cost (Appendix~\ref{sec:appendix_phase3_ablation}).

\section{Empirical evaluation}\label{sec:experiment}

We argue that generated text with causal-graph annotations that can serve as a substitute for real-world data should satisfy three desiderata: (1) the causal-graph annotations must be accurate, since they are expected to be used as ground-truth labels; (2) the text should be natural and difficult to distinguish from real-world text, in order to ensure the realism of text; and (3) the data should be practically usable for causal discovery algorithms applied to text, which is our ultimate engineering objective. Accordingly, we design three experiments to test whether the data generated by iTAG and by existing baselines satisfy these requirements.

\subsection{Experimental Setup}\label{sec:exp_setup}

\subsubsection{Baseline implementation}\label{sec:baselines}

As summarized in related work, text-with-graph generators fall into two common families (template-based vs.\ LLM-dependent); since within-family pipelines are similar, we instantiate one representative baseline per family, plus two LLM-dependent ablations that isolate Phase~2 design choices. The \textbf{template-based} baseline composes a paragraph by decomposing a target DAG into small causal motifs and instantiating hand-written templates, while the \textbf{LLM-dependent} baseline performs a single-shot graph-to-text generation from a serialized DAG over abstract node names (thus omitting Phase~2); on top of the same single-shot verbalization, \textbf{LLM-dependent+CA} adds a one-shot concept assignment step, whereas \textbf{LLM-dependent+ID} keeps single-shot concepts but adds an inverse-design-based revision loop during generation (Section~\ref{sec:inverse}).

Across all experiments, target graphs are sampled from the Phase~1 generator with $n\in\{3,\ldots,10\}$. For each fixed $n$, we draw the remaining Phase~1 structural parameters from Appendix~\ref{sec:appendix_itag_phase1_defaults} (Table~\ref{tab:phase1_param_space}). Appendix~\ref{sec:appendix_itag_phase1_defaults} sweeps this parameter space and finds that, after controlling for $n$, non-$n$ parameters have statistically non-significant effects (all corrected $p>0.05$; partial $\eta^2<0.01$), so we focus on $n$ as the main driver.

Unless otherwise specified, we use \texttt{claude-opus-4-1-20250805-thinking} as the default backbone LLM. All backbones are accessed via official public APIs, and we report exact API request templates for reproducibility (Appendix~\ref{sec:appendix_itag_repro}); we further repeat experiments with three additional backbones to test robustness (Appendix~\ref{sec:appendix_itag_robustness}). Unless otherwise noted, we use fixed API settings and set temperature $=0$ when applicable to maximize reproducibility rather than tune generation stochasticity. We generate 500 examples per method for each variable quantity, justified by a sample-size stability analysis showing that metrics converge well before this budget (Appendix~\ref{sec:appendix_sample_stability}).

\subsubsection{Real-world datasets}\label{sec:datasets}

We use three real-world corpora, medical MIMIC-IV-Note v2.2, business FinCausal 2025, and legal JUSTICE \cite{johnson2023mimicivnote, goldberger2000physionet, moreno2025financial, alali2021justice}, as natural-text references (Experiment~2) and as matched evaluation sets for transferability (Experiment~3). We sample 500 texts per corpus, construct a fixed concept (variable) set per text, and keep $n\in\{3,\dots,10\}$ to match the synthetic regime; Experiment~3 averages real-world metrics within each $n$-bucket (bucket sizes $N_n$ in Appendix~\ref{sec:appendix_bucket_sizes}; full corpus selection in Appendix~\ref{sec:appendix_data_anno}).

We obtain expert-consensus (\emph{silver-standard}) causal DAGs for the real corpora by majority vote over 11 trained annotators (blinded), who label only text-supported \emph{direct} edges under a DAG-only guideline; inter-annotator agreement is $\alpha=0.79$ (Krippendorff; guidelines and low-agreement adjudication in Appendices~\ref{sec:appendix_direct_edge_guidelines}--\ref{sec:appendix_low_agreement}).

\subsubsection{Experiment procedure and metrics}\label{sec:procedure&metrics}

All graph-based evaluations are performed on a fixed per-text concept set $\mathcal{V}$ (Appendix~\ref{sec:appendix_concept_set}); formal metric definitions and statistical tests are summarized in Appendix~\ref{sec:appendix_eval_metrics}.

\textbf{Experiment 1: Annotation accuracy of generated causal graphs.}
For each method, annotators re-annotate the direct-edge graph $\hat{G}$ from the generated text (blinded to the generation-time graph $G$), and we compare $\hat{G}$ against $G$ using edge-wise Precision/Recall and \textbf{graph-annotation $F1_{\mathrm{Ga}}$ ($\uparrow$)}, together with two structural distances (SHD/SID; $\downarrow$). Since SID is defined for DAGs, we compute SHD/SID after applying the deterministic DAG projection when majority-vote aggregation yields cycles (Appendix~\ref{sec:appendix_dag_projection}); we use SID purely as a direction-sensitive disagreement metric.

\textbf{Experiment 2: Naturalness and indistinguishability of generated text.}
We measure naturalness via binary ``real vs.\ generated'' classification, reporting \textbf{detectability $F1_{\mathrm{D}}$ ($\downarrow$)} for detecting the \emph{generated} class on balanced sets (random guessing yields $\approx 0.5$). We evaluate both (i) humans and (ii) four standard detectors (fastText, TextCNN, TSCNN, RoBERTa) trained under a fixed protocol \citep{joulin2017bag, kim2014convolutional, yang2018ts, liu2019roberta}; full training details and learning-curve stability are in Appendix~\ref{sec:appendix_detector_learning_curve}.

\textbf{Experiment 3: Transferability of causal discovery evaluation.}
We run representative text-based causal discovery algorithms on iTAG-generated corpora and on matched real-world corpora, evaluating them with \textbf{Graph $F1_{\mathrm{G}}$} ($\uparrow$) together with SHD ($\downarrow$) and SID ($\downarrow$), and test whether relative algorithm performance transfers. To factor out the shared monotonic dependence on graph size $n$, we center each metric within every $n$ bucket:
\[
\resizebox{\linewidth}{!}{$\displaystyle
\tilde{x}_{a,n} = x_{a,n} - \frac{1}{|\mathcal{A}|}\sum_{a'\in \mathcal{A}} x_{a',n}, \qquad
\tilde{y}_{a,n} = y_{a,n} - \frac{1}{|\mathcal{A}|}\sum_{a'\in \mathcal{A}} y_{a',n}.
$}
\]
and compute Pearson $r$, Spearman $\rho$, and linear-regression $R^2$ across the $|\mathcal{A}|\times 8$ algorithm--size pairs. Following prior surveys, we use RuleBayes, SCITE, and LLM-CG as representatives of statistical, supervised neural, and LLM-based paradigms \citep{asghar2016automatic, yang2022survey, sorgente2013automatic, li2021causality, antonucci2023zero}. For the LLM-CG baseline, we instantiate the LLM with \texttt{gpt-5-pro-2025-10-06}; we set temperature $=0$ for reproducibility, and leave other API settings at provider defaults.

\subsection{Main Results}\label{sec:results}

\subsubsection{Experiment 1: Annotation accuracy of generated causal graphs}\label{sec:exp1}

\begin{figure}[t]
    \centering
    \includegraphics[width=\columnwidth]{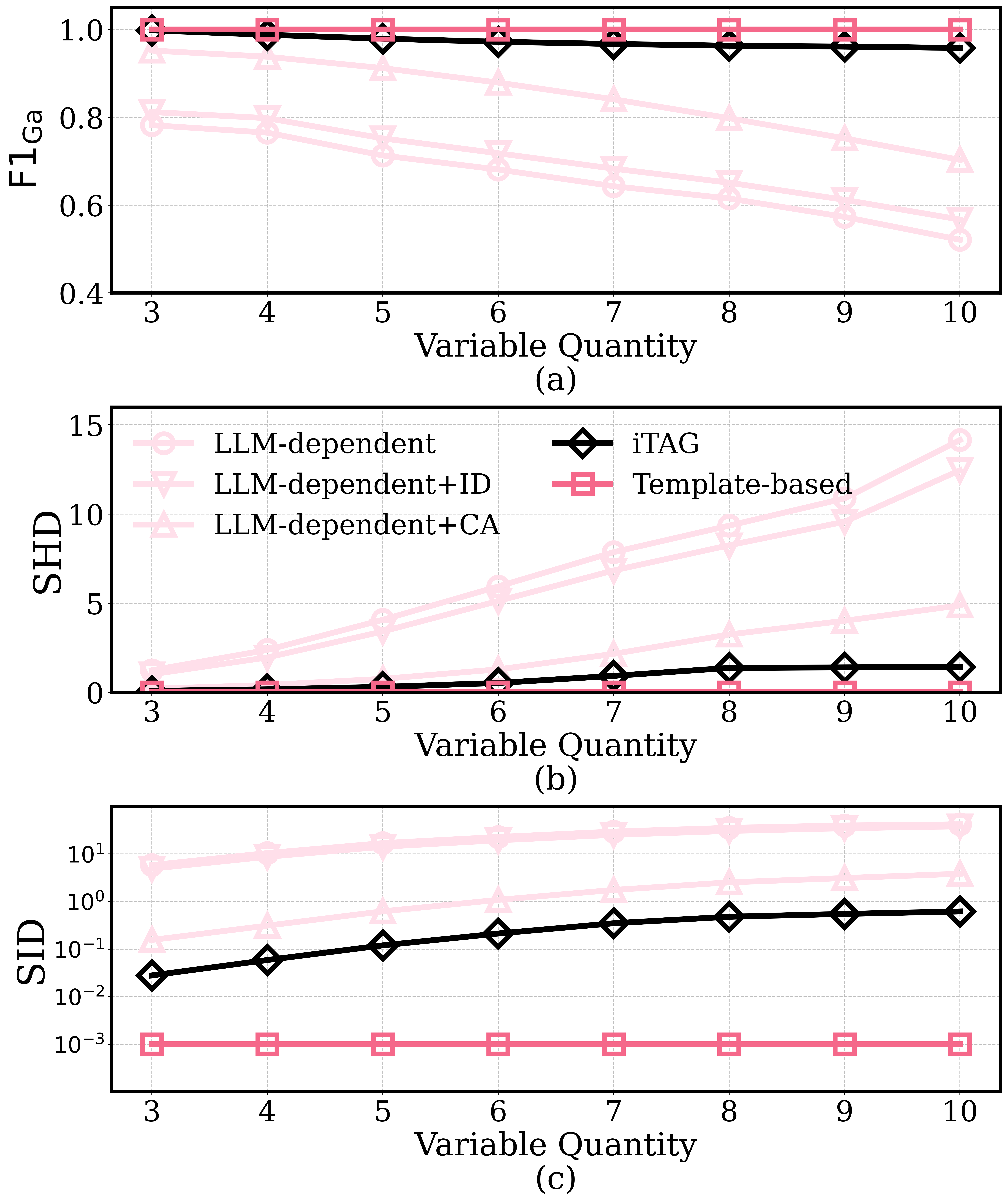}
    \caption{Annotation accuracy of generated causal graphs across methods on claude-opus-4-1.}
    \label{fig:mr1}
\end{figure}

Figure~\ref{fig:mr1} summarizes annotation accuracy as the variable quantity $n$ increases ($x$-axis, $n\in\{3,\dots,10\}$). Panel (a) reports graph-annotation $F1_{\mathrm{Ga}}$ ($\uparrow$), while panels (b--c) report SHD ($\downarrow$) and SID ($\downarrow$, log-scale; plotted as $\mathrm{SID}+\varepsilon$, $\varepsilon{=}10^{-3}$ to display perfect zeros). Each curve corresponds to one method. The template-based baseline serves as a sanity check: since its sentences are deterministically instantiated from the generation-time structure, the attached annotations match the generation-time graphs exactly ($F1_{\mathrm{Ga}} = 1$, $\mathrm{SHD}=\mathrm{SID}=0$ for all $n$), confirming that our evaluation protocol recognizes perfectly faithful annotations.

\begin{table*}[t]
    \centering
    \caption{Naturalness and indistinguishability of generated text across methods on claude-opus-4-1.}
    \label{tab:mr2}
    \begin{tabular}{lccccc}
    \toprule
    \textbf{Method} & \textbf{fastText} & \textbf{TextCNN} & \textbf{TSCNN} & \textbf{RoBERTa} & \textbf{Human} \\
    \midrule
    \rowcolor{gray!20}
    Template-based      & 0.98 & 0.97 & 0.96 & 0.99 & 0.81 \\
    LLM-dependent       & 0.58 & 0.61 & 0.59 & 0.64 & 0.57 \\
    LLM-dependent+CA    & 0.55 & 0.57 & 0.56 & 0.60 & 0.54 \\
    LLM-dependent+ID    & 0.56 & 0.59 & 0.58 & 0.62 & 0.55 \\
    \rowcolor{mypink}
    iTAG                & \textbf{0.52} & \textbf{0.54} & \textbf{0.53} & \textbf{0.57} & \textbf{0.51} \\
    \bottomrule
    \end{tabular}
\end{table*}

Against this oracle, iTAG stays near-perfect across graph sizes: $F1_{\mathrm{Ga}}\ge 0.95$ up to $n=10$, with only $\sim$one-edge SHD and mean $\mathrm{SID}<1$, implying that iTAG’s generation-time graphs remain reliable labels even as structural complexity grows. In contrast, the LLM-dependent baseline is strongly size-sensitive ($F1_{\mathrm{Ga}}$ drops from $\approx0.78$ to $\approx0.52$, and SHD rises into the low teens), indicating frequent omissions and spurious edges when an LLM verbalizes the entire graph in one pass.

The ablations attribute iTAG’s advantage primarily to Phase~2 concept substitution. One-shot concept assignment (LLM-dependent+CA) improves over LLM-dependent but still degrades with $n$ and trails iTAG, especially on SHD/SID, while generation-only inverse-design refinement (LLM-dependent+ID) yields only modest gains. To preempt the concern that aggregate metrics may hide non-edge violations ($a_{ij}=0$), Appendix~\ref{sec:appendix_error_decomp} decomposes errors into missed required edges vs.\ spurious edges on non-edges; iTAG’s remaining discrepancies are sparse (median spurious edges $<1$ across all $n$) and are dominated by borderline cases where annotators infer weak/indirect links.

Overall, Experiment~1 supports our first desideratum: iTAG produces texts whose attached causal graphs are trustworthy enough to be used as proxy ground-truth labels for downstream evaluation, maintaining high edge-wise accuracy and low structural disagreement as $n$ increases (with residual errors being sparse and largely attributable to borderline non-edge judgments; Appendix~\ref{sec:appendix_error_decomp}). Moreover, this result also clarifies the practical cost--benefit trade-off. iTAG uses more inference-time compute than the direct LLM-dependent baseline because it adds explicit verification and refinement, but this extra cost is modest and substantially reduces human correction effort. In our local logs, the baseline consumes 967.4 tokens per sample on average, whereas iTAG consumes 3684.2, i.e., about $3.8\times$ more. Notably, the baseline's annotation accuracy is relatively low and, for larger graphs, drops to around $0.5$---near random guessing---which substantially increases the burden of manual screening and correction. By spending this additional token budget to obtain much higher and more size-stable annotation accuracy, iTAG reduces the amount of human verification needed overall.

\textbf{Scalability.} The dominant verification set grows as $|R|=O(n^2)$ because ordered node pairs are checked under whole-graph direct-causality constraints, and we impose an explicit iteration cap. Empirically, however, the refinement loop typically converges in a small number of rounds (3--5), so within our tested regime the main practical issue is the token trade-off above rather than uncontrolled iteration growth.

\begin{figure*}[t]
    \centering
    \includegraphics[width=\textwidth]{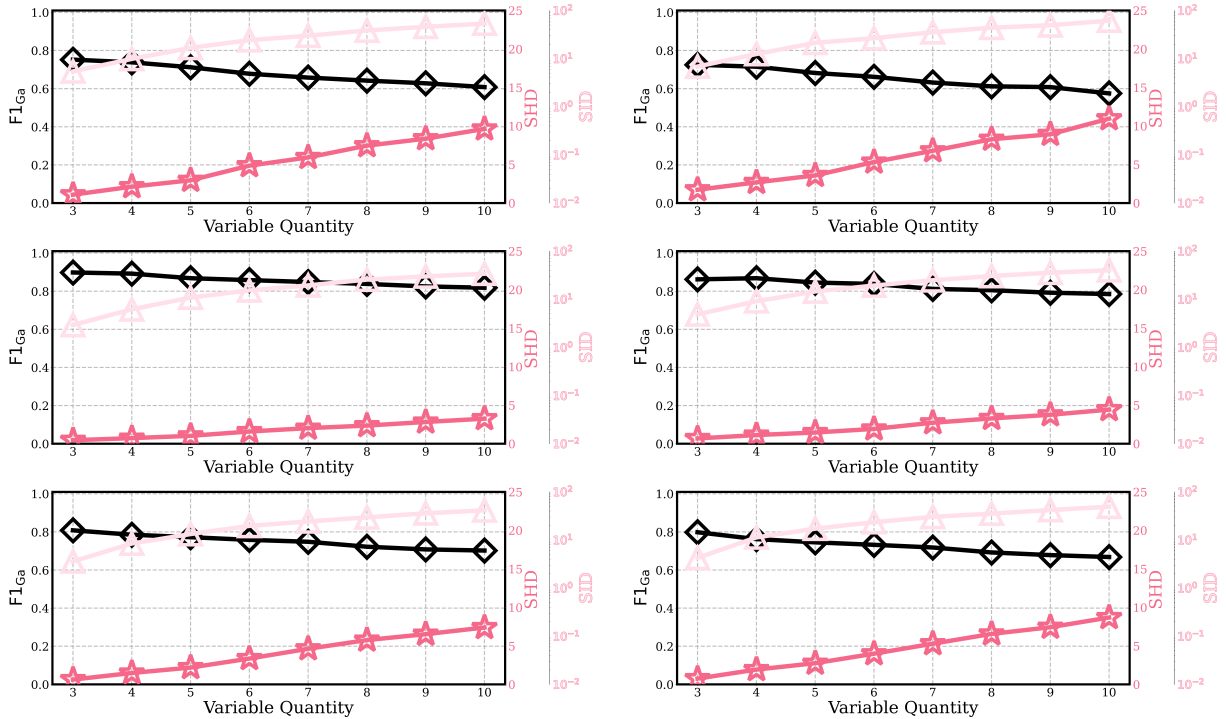}
    \caption{Transferability of causal discovery evaluation on claude-opus-4-1. Rows (top to bottom) are RuleBayes, SCITE, and LLM-CG; columns compare iTAG-generated corpora (left) vs.\ matched real-world corpora (right) across $n{=}3$--$10$. We report $F1_{\mathrm{G}}$ ($\uparrow$), SHD ($\downarrow$), and SID ($\downarrow$, log-scale; plotted as $\mathrm{SID}+\varepsilon$, $\varepsilon{=}10^{-3}$).}
    \label{fig:mr3}
\end{figure*}

\subsubsection{Experiment 2: Naturalness and indistinguishability of generated text}\label{sec:exp2}

\begin{table*}
  \centering
  \definecolor{mypink}{RGB}{224,80,128}
  \caption{Centered (within-$n$) agreement between algorithm scores on iTAG-generated vs.\ real-world corpora (24 algorithm--size pairs), controlling for the shared monotonic dependence on $n$. We report Pearson $r$ (two-sided $p$-values via a stratified permutation test that shuffles algorithm identities within each $n$-bucket; $B{=}10{,}000$ permutations), Spearman $\rho$ (two-sided), and linear-fit $R^2$ with 95\% CI.}
  \label{tab:mr3_stats}
  \resizebox{\textwidth}{!}{%
  \begin{tabular}{lccc}
    \toprule
    \multicolumn{1}{c}{\textbf{Pearson Corr.}} &
    \multicolumn{1}{c}{\textbf{Spearman Corr.}} &
    \multicolumn{1}{c}{\textbf{Linear Regr.}} \\
    \midrule
    \begin{tabular}{lcc}
      Metric & Corr. & p-value \\
      \midrule
      $F1_{\mathrm{G}}$  & 0.928 & $6.64\times 10^{-4}$ \\
      SHD & 0.927 & $7.69\times 10^{-4}$ \\
      SID & 0.921 & $1.78\times 10^{-3}$ \\
      \rowcolor{mypink!12}Average & \textbf{0.925} & / \\
    \end{tabular}
    &
    \begin{tabular}{lcc}
      Metric & Corr. & p-value \\
      \midrule
      $F1_{\mathrm{G}}$  & 0.926 & $8.81\times 10^{-4}$ \\
      SHD & 0.921 & $1.33\times 10^{-3}$ \\
      SID & 0.928 & $5.27\times 10^{-4}$ \\
      \rowcolor{mypink!12}Average & \textbf{0.925} & / \\
    \end{tabular}
    &
    \begin{tabular}{lcc}
      Metric & R$^2$ & 95\% CI \\
      \midrule
      $F1_{\mathrm{G}}$  & 0.861 & [0.703, 0.938] \\
      SHD & 0.859 & [0.699, 0.938] \\
      SID & 0.848 & [0.678, 0.932] \\
      \rowcolor{mypink!12}Average & \textbf{0.856} & [0.694, 0.936] \\
    \end{tabular} \\
    \bottomrule
  \end{tabular}%
  }
\end{table*}

Table~\ref{tab:mr2} reports, for each method (rows), the detectability $F1_{\mathrm{D}}$ of a binary ``real vs.\ generated'' detector (columns: four trained classifiers and humans) when predicting the \emph{generated} class on balanced test sets. Lower is better; in this setting, near-chance indistinguishability corresponds to $F1_{\mathrm{D}}\approx0.5$. All numbers are averaged over the three domains and all variable quantities $n\in\{3,\dots,10\}$.

The results show a clear ordering. Template-based text is easiest to spot: both humans ($F1_{\mathrm{D}}=0.81$) and all automatic detectors ($F1_{\mathrm{D}}\ge 0.96$) reliably identify it, indicating strong stylistic regularities despite perfect structural control. Direct LLM verbalization (LLM-dependent) is substantially harder to detect but remains nontrivially distinguishable (detectors $F1_{\mathrm{D}}=0.58$--$0.64$, humans $F1_{\mathrm{D}}=0.57$). The two ablations yield only incremental gains over this baseline: adding one-shot concept assignment (+CA) reduces detector/human $F1_{\mathrm{D}}$ by a few points, and adding refinement (+ID) yields similarly modest improvements.

Crucially, iTAG achieves the lowest detectability across \emph{all} evaluators (detectors $F1_{\mathrm{D}}=0.52$--$0.57$, humans $F1_{\mathrm{D}}=0.51$), i.e., closest to random guessing. Therefore, among the compared methods (holding the backbone LLM fixed), iTAG produces the most natural and least distinguishable text, supporting our second desideratum that the synthetic paragraphs can closely match real-corpus writing style rather than exhibiting easily exploitable artifacts. This conclusion is not driven by any particular graph size: the per-$n$ breakdown shows only small fluctuations and no monotonic degradation as $n$ increases (Appendix~\ref{sec:appendix_detection_by_n}).

\subsubsection{Experiment 3: Transferability of causal discovery evaluation}\label{sec:exp3}

Experiments~1 and~2 jointly show that iTAG is the only method that simultaneously satisfies our two primary desiderata for surrogate data: it achieves near-oracle annotation accuracy while producing text that is almost indistinguishable from real-world text for both humans and strong neural detectors. In contrast, the template-based baseline has perfect structural fidelity but highly unnatural, easily detected text, whereas LLM-dependent variants yield more natural text but substantially distorted causal graphs. Since neither family can provide realistic text with trustworthy graph annotations, our main transferability analysis focuses on iTAG.

Figure~\ref{fig:mr3} visualizes performance as graph size increases: each row is one algorithm, the left column uses iTAG-generated data and the right column uses real-world data; the $x$-axis is the variable quantity $n\in\{3,\ldots,10\}$, $F1_{\mathrm{G}}$ is read from the left $y$-axis, and SHD (linear) / SID (log) are read from the right $y$-axes. Points are corpus means.

From these curves, we make three increasingly strong inferences toward our transferability claim. 
\textbf{(i) Matched difficulty scaling.} For all three algorithms, increasing $n$ decreases $F1_{\mathrm{G}}$ and increases SHD/SID on \emph{both} iTAG-generated and real corpora, indicating that iTAG preserves how causal discovery becomes harder as graphs grow.
\textbf{(ii) Preserved algorithm comparisons.} The relative ordering is largely consistent: SCITE is best, LLM-CG is typically second, and RuleBayes is worst across $F1_{\mathrm{G}}$/SHD/SID, with only  near-ties at small $n$. This is the key property needed for benchmarking: conclusions about “which method is better” carry over.
\textbf{(iii) Limited absolute mismatch.} The generated--real gaps are small ($|\Delta{F1_{\mathrm{G}}}| \le 0.036$, $|\Delta \mathrm{SHD}| \le 1.33$, $|\Delta \mathrm{SID}| \le 8.14$ across all algorithm--$n$ points), suggesting iTAG acts as a closely matched (slightly easier) proxy.

A natural concern is that apparent alignment might be a trivial artifact of the shared monotonic dependence on $n$. To address this, we apply the within-$n$ centering protocol from Section~\ref{sec:procedure&metrics} and compute agreement across the $3\times 8=24$ algorithm--size pairs. Table~\ref{tab:mr3_stats} shows strong alignment after centering (Pearson $r\ge 0.921$, Spearman $\rho\ge 0.921$, all $p<0.001$; $R^2\ge 0.848$), meaning that algorithms that perform \emph{relatively} better on iTAG-generated data also perform \emph{relatively} better on real data at the same graph size. Appendix~\ref{sec:appendix_itag_robustness} further reports leave-one-algorithm-out and stratified bootstrap checks, guarding against the possibility that the correlation is driven by a single algorithm or sampling noise.

Overall, within our tested regime, evaluations conducted on iTAG-generated corpora are highly predictive of evaluations on matched real-world corpora, supporting our third desideratum that iTAG is practically usable for scalable benchmarking of text-based causal discovery (while not replacing real-world evaluation).

\section{Discussion}

We presented iTAG, which augments LLM-based text generation with an inverse-design concept assignment aligned to target causal graphs. Across three desiderata—annotation accuracy, naturalness, and transferability—we find that iTAG attains near-oracle structural accuracy, produces text that is difficult to distinguish from real texts, and yields causal-discovery metrics on synthetic corpora that closely track those on matched real corpora. 

For causal discovery from text, iTAG provides realistic corpora with trustworthy causal labels and predictable links to real-world performance, enabling systematic and low-cost benchmarking across graph sizes and structures without repeated large-scale annotation. It supports controlled comparisons and failure-mode analysis, helping narrow down algorithms that are likely to work in practice before investing in domain-specific data. iTAG is therefore a scalable evaluation aid rather than a substitute for real corpora.

\section*{Limitations}

iTAG cannot currently generate text with Structural Equation Model (SEM) annotations, i.e., explicit structural functions and effect parameters for each edge; it only provides adjacency-level causal graphs. This limitation arises because our Phase~1 sampler and Phase~2 concept assignment operate at the level of graph structure, and we deliberately avoid asking LLMs to specify numeric effect sizes or functional forms, which are much harder to validate reliably from text. As a result, iTAG is suitable for evaluating \emph{structural} causal discovery from text (recovering the presence or absence of edges), but cannot be used to benchmark methods that also estimate functional relationships or causal effect magnitudes. This does not affect our main goal: assessing text-based causal structure discovery, because all evaluated algorithms focus on edge-level graphs. A natural extension would be to sample full SCMs for each DAG from a controlled family (e.g., linear or nonlinear SEMs), simulate data to obtain ground-truth parameters, and then encode qualitative or quantitative cues about effect strength into the generated text; such a hybrid design would preserve the current structural guarantees while enabling parameter-level evaluation.

A second limitation is that we only instantiate and validate iTAG on relatively small graphs (3–10 variables) and three English domains (business, medical, legal). These choices are driven by the cost of multi-annotator validation and by the observation that typical decision-making narratives rarely involve more than a handful of key variables. However, this means our experiments do not fully cover very large, document-level causal structures or other languages and genres, and text-based causal discovery methods might face additional challenges in those settings. While this constraint does not undermine our core claims about structure-level fidelity, naturalness, and transferability within the tested regime, future work should scale iTAG to larger or hierarchical graphs and more domains or languages; moreover, because iTAG can generate highly natural synthetic text, we recommend using it for algorithmic benchmarking and controlled stress tests with appropriate disclosure/safety filtering when releasing corpora.

A further limitation concerns negative edges $a_{ij}=0$. Our inverse-design loop explicitly reasons about both existing and non-existing relations, but deciding from text that no direct causal relation holds between two variables is inherently harder than confirming the presence of a relation. Short narratives may underspecify potential confounders or indirect pathways, and our LLM-based verifier can still misclassify borderline cases. Although Section \ref{sec:exp1} shows that spurious edges are rare under our evaluation protocol, future work should incorporate more principled uncertainty modeling or calibration for non-edge predictions, and explore adversarial test cases that specifically target these weaknesses.

\section*{Acknowledgments}
We thank the reviewers and area chairs for their constructive feedback. We used LLMs only for linguistic and presentational assistance, including terminology checking, grammar correction, translation from the authors' native language to English, and improving sentence flow while preserving the original scientific meaning. They were not used for research ideation, experimental design, data analysis, or scientific content generation; all scientific insights, methodological decisions, and conclusions are the authors' own.


\bibliography{custom}

@inproceedings{shrestha2022automatically,
  title={Automatically explaining a model: Using deep neural networks to generate text from causal maps},
  author={Shrestha, Anish and Mielke, Kyle and Nguyen, Tuong Anh and Giabbanelli, Philippe J},
  booktitle={2022 Winter simulation conference (WSC)},
  pages={2629--2640},
  year={2022},
  organization={IEEE}
}

@article{phatak2024narrating,
  title={Narrating causal graphs with large language models},
  author={Phatak, Atharva and Mago, Vijay K and Agrawal, Ameeta and Inbasekaran, Aravind and Giabbanelli, Philippe J},
  journal={arXiv preprint arXiv:2403.07118},
  year={2024}
}

@inbook{Gandee,
author = {Gandee, Tyler and Giabbanelli, Philippe},
year = {2024},
month = {10},
pages = {359-376},
title = {Combining Natural Language Generation and Graph Algorithms to Explain Causal Maps Through Meaningful Paragraphs},
isbn = {978-3-031-75598-9},
doi = {10.1007/978-3-031-75599-6_25}
}

@article{gandee2025faithful,
  title={Faithful Narratives from Complex Conceptual Models: Should Modelers or Large Language Models Simplify Causal Maps?},
  author={Gandee, Tyler J and Giabbanelli, Philippe J},
  journal={Machine Learning and Knowledge Extraction},
  volume={7},
  number={4},
  pages={116},
  year={2025},
  publisher={MDPI}
}

@article{stuart2010inverse,
  title={Inverse problems: a Bayesian perspective},
  author={Stuart, Andrew M},
  journal={Acta numerica},
  volume={19},
  pages={451--559},
  year={2010},
  publisher={Cambridge University Press}
}

@inproceedings{pfaff2020learning,
  title={Learning mesh-based simulation with graph networks},
  author={Pfaff, Tobias and Fortunato, Meire and Sanchez-Gonzalez, Alvaro and Battaglia, Peter},
  booktitle={International conference on learning representations},
  year={2020}
}

@inproceedings{kim2019deep,
  title={Deep fluids: A generative network for parameterized fluid simulations},
  author={Kim, Byungsoo and Azevedo, Vinicius C and Thuerey, Nils and Kim, Theodore and Gross, Markus and Solenthaler, Barbara},
  booktitle={Computer graphics forum},
  volume={38},
  number={2},
  pages={59--70},
  year={2019},
  organization={Wiley Online Library}
}

@misc{wei2023chainofthoughtpromptingelicitsreasoning,
      title={Chain-of-Thought Prompting Elicits Reasoning in Large Language Models}, 
      author={Jason Wei and Xuezhi Wang and Dale Schuurmans and Maarten Bosma and Brian Ichter and Fei Xia and Ed Chi and Quoc Le and Denny Zhou},
      year={2023},
      eprint={2201.11903},
      archivePrefix={arXiv},
      primaryClass={cs.CL},
      url={https://arxiv.org/abs/2201.11903}, 
}

@article{wang2022self,
  title={Self-consistency improves chain of thought reasoning in language models},
  author={Wang, Xuezhi and Wei, Jason and Schuurmans, Dale and Le, Quoc and Chi, Ed and Narang, Sharan and Chowdhery, Aakanksha and Zhou, Denny},
  journal={arXiv preprint arXiv:2203.11171},
  year={2022}
}

@article{yao2023tree,
  title={Tree of thoughts: Deliberate problem solving with large language models},
  author={Yao, Shunyu and Yu, Dian and Zhao, Jeffrey and Shafran, Izhak and Griffiths, Tom and Cao, Yuan and Narasimhan, Karthik},
  journal={Advances in neural information processing systems},
  volume={36},
  pages={11809--11822},
  year={2023}
}

@misc{zhou2023leasttomostpromptingenablescomplex,
      title={Least-to-Most Prompting Enables Complex Reasoning in Large Language Models}, 
      author={Denny Zhou and Nathanael Schärli and Le Hou and Jason Wei and Nathan Scales and Xuezhi Wang and Dale Schuurmans and Claire Cui and Olivier Bousquet and Quoc Le and Ed Chi},
      year={2023},
      eprint={2205.10625},
      archivePrefix={arXiv},
      primaryClass={cs.AI},
      url={https://arxiv.org/abs/2205.10625}, 
}

@article{erdds1959random,
  title={On random graphs I},
  author={Erd{\H{o}}s, Paul and R{\'e}nyi, Alfr{\'e}d},
  journal={Publ. math. debrecen},
  volume={6},
  number={290-297},
  pages={18},
  year={1959}
}

@article{erd6s1960evolution,
  title={On the evolution of random graphs},
  author={Erd{\H{o}}s, Paul and R{\'e}nyi, Alfr{\'e}d},
  journal={Publ. Math. Inst. Hungar. Acad. Sci},
  volume={5},
  pages={17--61},
  year={1960}
}

@misc{johnson2023mimicivnote,
  author       = {Johnson, A. and Pollard, T. and Horng, S. and Celi, L. A. and Mark, R.},
  title        = {{MIMIC-IV-Note: Deidentified free-text clinical notes} (version 2.2)},
  year         = {2023},
  publisher    = {PhysioNet},
  doi          = {10.13026/1n74-ne17},
  note         = {RRID:SCR\_007345}
}

@article{goldberger2000physionet,
  author       = {Goldberger, A. and Amaral, L. and Glass, L. and Hausdorff, J. and Ivanov, P. C. and Mark, R. and Stanley, H. E.},
  title        = {{PhysioBank, PhysioToolkit, and PhysioNet: Components of a new research resource for complex physiologic signals}},
  journal      = {Circulation [Online]},
  volume       = {101},
  number       = {23},
  pages        = {e215--e220},
  year         = {2000},
  note         = {RRID:SCR\_007345}
}

@inproceedings{moreno2025financial,
  title={The financial document causality detection shared task (FinCausal 2025)},
  author={Moreno-Sandoval, Antonio and Zamorano, Jordi Porta and Carbajo-Coronado, Blanca and Torterolo, Yanco and Samy, Doaa},
  booktitle={Proceedings of the Joint Workshop of the 9th Financial Technology and Natural Language Processing (FinNLP), the 6th Financial Narrative Processing (FNP), and the 1st Workshop on Large Language Models for Finance and Legal (LLMFinLegal)},
  pages={214--221},
  year={2025}
}

@article{alali2021justice,
  title={Justice: A benchmark dataset for supreme court's judgment prediction},
  author={Alali, Mohammad and Syed, Shaayan and Alsayed, Mohammed and Patel, Smit and Bodala, Hemanth},
  journal={arXiv preprint arXiv:2112.03414},
  year={2021}
}

@article{asghar2016automatic,
  title={Automatic extraction of causal relations from natural language texts: a comprehensive survey},
  author={Asghar, Nabiha},
  journal={arXiv preprint arXiv:1605.07895},
  year={2016}
}

@article{yang2022survey,
  title={A survey on extraction of causal relations from natural language text},
  author={Yang, Jie and Han, Soyeon Caren and Poon, Josiah},
  journal={Knowledge and Information Systems},
  volume={64},
  number={5},
  pages={1161--1186},
  year={2022},
  publisher={Springer}
}

@article{sorgente2013automatic,
  title={Automatic extraction of cause-effect relations in Natural Language Text.},
  author={Sorgente, Antonio and Vettigli, Giuseppe and Mele, Francesco},
  journal={DART@ AI* IA},
  volume={2013},
  pages={37--48},
  year={2013}
}

@article{li2021causality,
  title={Causality extraction based on self-attentive BiLSTM-CRF with transferred embeddings},
  author={Li, Zhaoning and Li, Qi and Zou, Xiaotian and Ren, Jiangtao},
  journal={Neurocomputing},
  volume={423},
  pages={207--219},
  year={2021},
  publisher={Elsevier}
}

@article{antonucci2023zero,
  title={Zero-shot causal graph extrapolation from text via llms},
  author={Antonucci, Alessandro and Piqu{\'e}, Gregorio and Zaffalon, Marco},
  journal={arXiv preprint arXiv:2312.14670},
  year={2023}
}

@inproceedings{joulin2017bag,
  title={Bag of tricks for efficient text classification},
  author={Joulin, Armand and Grave, Edouard and Bojanowski, Piotr and Mikolov, Tom{\'a}{\v{s}}},
  booktitle={Proceedings of the 15th conference of the European chapter of the association for computational linguistics: volume 2, short papers},
  pages={427--431},
  year={2017}
}

@article{kim2014convolutional,
  title={Convolutional neural networks for sentence classification},
  author={Kim, Yoon},
  journal={arXiv preprint arXiv:1408.5882},
  year={2014}
}

@article{liu2019roberta,
  title={Roberta: A robustly optimized bert pretraining approach},
  author={Liu, Yinhan and Ott, Myle and Goyal, Naman and Du, Jingfei and Joshi, Mandar and Chen, Danqi and Levy, Omer and Lewis, Mike and Zettlemoyer, Luke and Stoyanov, Veselin},
  journal={arXiv preprint arXiv:1907.11692},
  year={2019}
}

@article{yang2018ts,
  title={TS-CNN: Text steganalysis from semantic space based on convolutional neural network},
  author={Yang, Zhongliang and Wei, Nan and Sheng, Junyi and Huang, Yongfeng and Zhang, Yu-Jin},
  journal={arXiv preprint arXiv:1810.08136},
  year={2018}
}

@inproceedings{faller2024self,
  title={Self-compatibility: Evaluating causal discovery without ground truth},
  author={Faller, Philipp M and Vankadara, Leena C and Mastakouri, Atalanti A and Locatello, Francesco and Janzing, Dominik},
  booktitle={International Conference on Artificial Intelligence and Statistics},
  pages={4132--4140},
  year={2024},
  organization={PMLR}
}

@article{hiremath2024losam,
  title={LoSAM: Local Search in Additive Noise Models with Unmeasured Confounders, a Top-Down Global Discovery Approach},
  author={Hiremath, Sujai and Gan, Kyra and Ghosal, Promit},
  journal={arXiv e-prints},
  pages={arXiv--2410},
  year={2024}
}

@article{brouillard2024landscape,
  title={The landscape of causal discovery data: Grounding causal discovery in real-world applications},
  author={Brouillard, Philippe and Squires, Chandler and Wahl, Jonas and Kording, Konrad P and Sachs, Karen and Drouin, Alexandre and Sridhar, Dhanya},
  journal={arXiv preprint arXiv:2412.01953},
  year={2024}
}

@inproceedings{ding2025multi,
  title={A Multi-Level Benchmark for Causal Language Understanding in Social Media Discourse},
  author={Ding, Xiaohan and Ping, Kaike and {\c{C}}ar{\i}k, Buse and Rho, Eugenia},
  booktitle={Proceedings of the 2025 Conference on Empirical Methods in Natural Language Processing},
  pages={28764--28778},
  year={2025}
}

@article{wang2025causalenhance,
  title={CausalEnhance: Knowledge-Enhanced Pre-training for Causality Identification and Extraction},
  author={Wang, Meiyun and Izumi, Kiyoshi and Sakaji, Hiroki},
  journal={Knowledge-Based Systems},
  pages={114447},
  year={2025},
  publisher={Elsevier}
}

@inproceedings{gujarathi2022study,
  title={A study of extracting causal relationships from text},
  author={Gujarathi, Pranav and Reddy, Manohar and Tayade, Neha and Chakraborty, Sunandan},
  booktitle={Proceedings of SAI Intelligent Systems Conference},
  pages={807--828},
  year={2022},
  organization={Springer}
}

\newpage

\appendix

\section*{Contents of Appendices}
\vspace{1em}
\noindent
\hyperref[sec:prompt]{A \quad Prompt Templates
    \dotfill \pageref{sec:prompt}}\\[0.5em]
\hyperref[sec:appendix_itag_impl]{B \quad iTAG Method and Implementation Details
    \dotfill \pageref{sec:appendix_itag_impl}}\\[0.5em]
\hyperref[sec:appendix_supp]{C \quad Supplementary Experiments and Analyses
    \dotfill \pageref{sec:appendix_supp}}\\[0.5em]
\hyperref[sec:appendix_data_anno]{D \quad Real-World Datasets and Human Annotation Protocol
    \dotfill \pageref{sec:appendix_data_anno}}
\vspace{2em}

\newpage

\section{Prompt Templates}\label{sec:prompt}

\begin{tcolorbox}[colback=mypink, colframe=myframepink, title=\textbf{Phase 2 of iTAG}, breakable]
Adjacency Matrix:\\
{[}Matrix{]}
\\
\\
====================
\\
\\
Task: Please assign concepts from a meaningful real-world [domain/series of events] to the [N] nodes in the causal DAG represented by this adjacency matrix, such that (i) all relationships represented by 1s are plausible \emph{direct} causal relations between the assigned concepts, and (ii) relationships represented by 0s do \emph{not} imply any \emph{direct} causal relation (they may be unrelated, indirect, or confounded).
\\
\\
Requirements: In your thinking, please use the following separators to assist your reasoning, but \emph{do not} print these separators or intermediate steps in the output; only output the final result when you are satisfied with it:\\
----Let me first analyze carefully---- \\
(First list all relationships between nodes represented by 1s in the matrix and all non-existent relationships represented by 0s in the matrix)\\
----First attempt---- \\
(Then write out the concepts corresponding to the nodes)\\
----Check for errors---- \\
(Please use the complete paradigm '''First, imagine that in the real world, [variable A] occurs (or takes some value) and [variable B] subsequently occurs (or takes some value). If [variable A] had not occurred (or had taken a different value), would [variable B] still occur in the same way (or maintain the same value) under the same background conditions? If in the counterfactual scenario where [variable A] did not occur, [variable B] significantly changes (either does not occur at all, or occurs in a substantially different way, time, intensity, or characteristics), and this change is systematic rather than accidental, while all other potential background conditions and common causes that might affect [variable B] remain constant, then we can reasonably infer a causal relationship between [variable A] and [variable B], meaning [variable A] is a cause of [variable B]. Conversely, if in the counterfactual scenario, even when [variable A] does not occur, [variable B] still occurs in essentially the same way, or changes in [variable B] can be fully explained by changes in other variables, and this situation stably repeats across various background conditions, this indicates there is no direct or substantial causal relationship between [variable A] and [variable B], and the observed correlation between them may be coincidental, a spurious association due to common causes, or an indirect effect mediated through other variables rather than a true causal connection.''' to check whether the concepts are compatible with relationships marked by 1s in the matrix and do not systematically support relationships marked by 0s as direct causes or effects. If causal relationships are unreasonable, consider the reasons for errors and avoid them in the next attempt)\\
Begin second analysis\\
----Second attempt---- \\
...
\\
\\
Your answer should be in JSON format:
\begin{verbatim}
{
  "Existing causal relationships (
  values of 1 in the matrix)": [
    "Node 0 → Node 1",
    ...
  ],
  "Non-existing causal relationships 
  (values of 0 in the matrix)": [
    "Node 0 → Node 1",
    ...
  ],
  "Real concepts assigned to variables
  ": [
    "Node 0: ___",
    ...
  ],
  "Relationship verification": {
    "Existing causal relationships": [
      "___ (natural language 
      description conforming to the 
      reasoning paradigm)",
      ...
    ],
    "Non-existing causal relationships
    ": [
      "___ (natural language 
      description conforming to the 
      reasoning paradigm)",
      ...
    ]
  }
}
\end{verbatim}
\end{tcolorbox}

\begin{tcolorbox}[colback=mypink, colframe=myframepink, title=\textbf{Component 3 of iTAG}, breakable]
Concepts:\\
{[}Concepts{]}
\\
\\
Adjacency matrix between concepts:\\
{[}Adjacency Matrix{]}
\\
\\
====================
\\
\\
Task: Please express all concepts clearly in a paragraph of natural language (implicitly conveying relationships between concepts rather than explicitly stating them), without introducing any additional concepts.
\\
\\
Requirements: Before producing your final answer, silently go through the following steps in your own reasoning, but \emph{do not} print these steps in the output: (i) identify which concept pairs have a direct causal relationship (entries of 1 in the matrix) and which do not (entries of 0); (ii) plan a coherent paragraph in which all concepts appear and \emph{each} relationship marked as 1 is naturally suggested by the narrative at least once; and (iii) check that the paragraph does not explicitly or implicitly assert any direct causal relationship for pairs marked as 0, and that no additional concepts are introduced. After this internal check, output only the final result in the specified JSON format.
\\
\\
Your answer should be in JSON format:
\begin{verbatim}
{
  "Natural language description": "..."
}
\end{verbatim}
\end{tcolorbox}

\begin{tcolorbox}[colback=mypink, colframe=myframepink, title=\textbf{LLM causal discovery prompt}, breakable]
Text:\\
{[}Text{]}
\\
\\
Important concepts appearing in the text:\\
{[}Important concepts{]}
\\
\\
====================
\\
\\
Task: For the text and the important concepts appearing in it, please infer the **direct causal relationships** between each concept based on the text and common sense reasoning (causal relationships are not the same as correlations. For example, high temperature has causal relationships with both the number of drownings and ice cream sales, but the number of drownings and ice cream sales only have correlation without direct causal relationship).\\
\\
Requirements: The format for annotating causal relationships for each text should be:\\
0101 (means that the first concept has direct causal relationships with the second and fourth concepts, and the first concept is the cause of the second and fourth concepts)
0010 (means that the second concept has a direct causal relationship with the third concept, and the second concept is the cause of the third concept)
0000 (means that the third concept is not the cause of any other concept)
0100 (similarly, ...)\\
\\
Your response must be in JSON format containing the following:
\begin{verbatim}
{
  "adjacency matrix": [
    [0,1,0,...],
    [0,0,1,...],
    ...
  ]
}
\end{verbatim}
\end{tcolorbox}

\section{iTAG Method and Implementation Details}\label{sec:appendix_itag_impl}

This appendix provides additional implementation details required to reproduce iTAG
(Sections~\ref{sec:graph}--\ref{sec:text} and Algorithm~\ref{alg:concept_substitution}).
It covers (i) the domain-scoped concept schema and the initialization procedure used in Phase~2
(Appendix~\ref{sec:appendix_itag_schema_init});
(ii) the full Phase~1 structural input space, wide-range sensitivity analysis, and the sampling protocol
used in our reported experiments (Appendix~\ref{sec:appendix_itag_phase1_defaults});
(iii) the CounterfactualVerification protocol, self-consistency aggregation, and the diagnostic mismatch
objective $L_b$ used by the propose--evaluate--refine loop in Phase~2 (Appendix~\ref{sec:appendix_itag_verifier});
and (iv) backbone LLM identifiers, decoding parameters, and API request templates used in our experiments
(Appendix~\ref{sec:appendix_itag_repro}), as well as robustness and statistical-stability checks used in
Experiment~3 (Appendix~\ref{sec:appendix_itag_robustness} and Appendix~\ref{sec:appendix_sample_stability}).
Prompt texts for concept assignment and textual transformation are given in Appendix~\ref{sec:prompt}.

\subsection{Domain-scoped concept schema and Phase~2 initialization}\label{sec:appendix_itag_schema_init}

\subsubsection{Domain-scoped concept schema}\label{sec:appendix_itag_schema}
Phase~2 searches for a concept assignment $C=(c_1,\dots,c_n)$ from an admissible concept set $\mathcal{C}$ that is constrained by
a \emph{domain-scoped concept schema}. The goal of the schema is to make (i) causal verification from short narratives feasible,
and (ii) the subsequent Phase~3 text generation stable, by preventing degenerate concept choices (e.g., overlapping synonyms,
un-intervenable abstractions, or highly imbalanced granularity).

We enforce three constraints:

\paragraph{(1) Non-overlap.}
Distinct nodes must be assigned \emph{semantically distinct} concepts.
Formally, for any $i\neq j$, $c_i$ and $c_j$ should not be synonyms, near-synonyms, or trivially nested concepts
(e.g., ``\emph{skill}'' vs.\ ``\emph{ability}'', or ``\emph{court ruling}'' vs.\ ``\emph{legal decision}'').
This constraint reduces ambiguity when humans and models infer direct edges, and prevents spurious edges caused by conceptual redundancy.

\paragraph{(2) Comparable granularity.}
All concepts should be at a comparable level of abstraction within the chosen domain.
For example, a valid business assignment might contain entities such as ``\emph{marketing budget}'', ``\emph{customer demand}'',
and ``\emph{revenue}'', rather than mixing ``\emph{macroeconomic policy}'' with ``\emph{an individual customer's click}''.
This constraint improves interpretability and reduces systematic confounds where one overly broad concept becomes a plausible
cause of many others.

\paragraph{(3) Intervention-meaningfulness.}
Each concept must admit a meaningful counterfactual intervention in the sense used by our counterfactual-style verifier:
it should be possible to imagine ``\emph{concept $c_i$ occurs/takes a higher value}'' versus ``\emph{$c_i$ does not occur/takes a lower value}''
under comparable background conditions.
Concepts that are purely definitional, tautological, or ill-posed for interventions (e.g., ``\emph{fate}'', ``\emph{the truth}'') are disallowed.
This constraint is critical because CounterfactualVerification relies on counterfactual reasoning (Section~\ref{sec:inverse}) and degrades
when concepts do not support stable ``do''-style mental interventions.

\subsubsection{How the schema is enforced}\label{sec:appendix_itag_schema_enforce}
We enforce the schema via a combination of prompt-level constraints (Appendix~\ref{sec:prompt}) and lightweight validators:

\begin{itemize}
    \item \textbf{Prompt constraints.} In Phase~2, the prompt explicitly instructs the LLM to (i) use a single coherent real-world domain,
    (ii) assign distinct concepts to all nodes, and (iii) verify edge and non-edge plausibility via a counterfactual paradigm. In Phase~3,
    the prompt forbids introducing any additional concepts beyond the assigned set.
    \item \textbf{String-level checks (non-overlap).} We reject exact duplicates and near-duplicates after normalization
    (lowercasing, removing punctuation, stripping determiners). When a violation is detected, we request a minimal revision that replaces
    only the offending concept(s).
    \item \textbf{LLM-based schema check (optional).} For rare borderline cases (e.g., potential synonymy or questionable intervention meaning),
    we query the verifier with a short schema-check prompt that asks whether two concept strings are distinct and intervenable.
    This check is used as a tie-breaker and is not required for every sample.
\end{itemize}

\subsubsection{InitialConceptAssignment}\label{sec:appendix_itag_initial_concepts}
Algorithm~\ref{alg:concept_substitution} begins with \texttt{InitialConceptAssignment}, which produces an initial concept assignment $C^{(0)}$ that \emph{roughly}
conforms to the target adjacency matrix $A$.

\paragraph{Inputs.}
The initializer takes as input (i) the adjacency matrix $A$, and (ii) a domain label (business/medical/legal in our experiments).
The domain label is fixed per corpus to match the corresponding real-world domain used in Experiments~2--3.

\paragraph{Procedure.}
Concepts are proposed by prompting the backbone LLM with the Phase~2 concept-assignment template (Appendix~\ref{sec:prompt}, Component~2),
which requires the model to (i) enumerate required edges ($a_{ij}=1$) and non-edges ($a_{ij}=0$), and (ii) assign a set of
domain-consistent concepts while checking plausibility under the counterfactual paradigm.
If a schema validator flags overlap or non-intervenability, we apply a single minimal revision step that replaces only the flagged concepts.

\paragraph{Output.}
The output is an assignment $C^{(0)}$ and a structured list of relationships
$\mathcal{R}=\{(i,j):a_{ij}=1\}\cup\{(i,j):a_{ij}=0,i\neq j\}$ produced by \texttt{AnalyzeCausalStructure}.
This relationship list is passed to \texttt{CounterfactualVerification} in each iteration of the refinement loop.

\subsection{Phase~1 structural parameters: full input space, sensitivity analysis, and sampling protocol}
\label{sec:appendix_itag_phase1_defaults}

Across all experiments we vary the variable quantity $n \in \{3,\ldots,10\}$.
To avoid dependence on a small set of hand-picked structural defaults, we (i) explicitly characterize the full set
of Phase~1 structural inputs, (ii) run a wide-range one-factor sensitivity analysis over these parameters,
and (iii) in the main experiments, draw non-$n$ structural parameters from uniform distributions over the tested ranges.
This appendix reports the tested parameter space, the sensitivity results, and the resulting sampling protocol.

\paragraph{Phase~1 inputs and admissible domains.}
Phase~1 (Section~\ref{sec:graph}) samples a DAG using the following control parameters:
variable quantity $n$, expected density $p$, degree limits (\textit{max\_parents}, \textit{max\_children}),
and motif controls (confounder ratio $\gamma_c$, collider ratio $\gamma_v$, mediator-chain count $\lambda$).
Their admissible domains are:
$p \in (0,1)$,
$\textit{max\_parents},\textit{max\_children} \in \{0,\ldots,n-1\}$,
$\gamma_c,\gamma_v \in [0,1]$,
and $\lambda \in \{0,\ldots,n-2\}$.

\paragraph{Tested parameter space and main-experiment sampling.}
Table~\ref{tab:phase1_param_space} summarizes the wide tested ranges and the sampling distributions used in reported
experiments. For parameters whose feasible set depends on $n$ (degree limits and $\lambda$), sampling is conditional on the chosen $n$.

\begin{table*}[t]
\centering
\resizebox{\textwidth}{!}{
\setlength{\tabcolsep}{4pt}
\begin{tabular}{lccc}
\hline
\textbf{Parameter} & \textbf{Admissible domain} & \textbf{Tested range (wide)} & \textbf{Main sampling} \\
\hline
Variable quantity $n$ &
$\{3,\ldots,10\}$ &
$\{3,\ldots,10\}$ &
$\mathrm{Unif}\{3,\ldots,10\}$ \\
Expected density $p$ &
$(0,1)$ &
$[0.05,\,0.80]$ &
$\mathrm{Unif}(0.05,0.80)$ \\
\textit{max\_parents} &
$\{0,\ldots,n-1\}$ &
$\{1,\ldots,n-1\}$ &
$\mathrm{Unif}\{1,\ldots,n-1\}$ \\
\textit{max\_children} &
$\{0,\ldots,n-1\}$ &
$\{1,\ldots,n-1\}$ &
$\mathrm{Unif}\{1,\ldots,n-1\}$ \\
Confounder ratio $\gamma_c$ &
$[0,1]$ &
$[0.00,\,0.80]$ &
$\mathrm{Unif}(0.00,0.80)$ \\
Collider ratio $\gamma_v$ &
$[0,1]$ &
$[0.00,\,0.80]$ &
$\mathrm{Unif}(0.00,0.80)$ \\
Mediator chains $\lambda$ &
$\{0,\ldots,n-2\}$ &
$\{0,\ldots,n-2\}$ &
$\mathrm{Unif}\{0,\ldots,n-2\}$ \\
\hline
\end{tabular}
}
\caption{
Phase~1 structural parameter domains, wide tested ranges, and the sampling distributions used in reported experiments.
Aggregated metrics that pool over $n$ use an equal number of samples per $n$ (equivalently, a uniform mixture over $n$).
}
\label{tab:phase1_param_space}
\end{table*}

\paragraph{One-factor sensitivity analysis (wide ranges).}
For each Phase~1 parameter $\theta \in \{p,\textit{max\_parents},\textit{max\_children},\gamma_c,\gamma_v,\lambda\}$,
we run a one-factor-at-a-time sweep over the tested range in Table~\ref{tab:phase1_param_space}, while holding the other
Phase~1 parameters fixed at a moderate reference configuration.\footnote{We use the same reference configuration across $n$
whenever feasible; degree-limit and $\lambda$ sweeps are performed over the full feasible integer set given each $n$.}
For each sweep value and each $n\in\{3,\ldots,10\}$, we generate a pilot batch of graphs and run the full iTAG pipeline,
evaluating the downstream metrics used in Experiments~1--2:
graph-annotation $F1_{\mathrm{Ga}}$ (higher is better), SHD/SID (lower is better), and detectability $F1_{\mathrm{D}}$ (lower is better).
In Table~\ref{tab:phase1_sensitivity_summary}, each ``range'' reports the minimum and maximum of the \emph{mean} metric across the sweep grid
(after averaging over the three domains and pooling $n=3$--$10$). For $F1_{\mathrm{D}}$, we report the macro-average across the four trained detectors
and humans (same protocol as Experiment~2; Section~\ref{sec:procedure&metrics}).

\begin{table*}[t]
\centering
\resizebox{\textwidth}{!}{
\setlength{\tabcolsep}{4pt}
\begin{tabular}{lcccccp{2.3cm}}
\hline
\textbf{Swept parameter} &
\textbf{Grid size} &
$\mathbf{F1_{\mathrm{Ga}}}$ \textbf{range} &
\textbf{SHD range} &
\textbf{SID range} &
$\mathbf{F1_{\mathrm{D}}}$ \textbf{range} &
\textbf{Conclusion} \\
\hline
$p$ &
$11$ &
$[0.948,\,0.971]$ &
$[0.86,\,1.24]$ &
$[0.42,\,0.68]$ &
$[0.508,\,0.536]$ &
small vs.\ changing $n$ \\
\textit{max\_parents} &
$|\{1,\ldots,n-1\}|$ &
$[0.952,\,0.968]$ &
$[0.91,\,1.18]$ &
$[0.46,\,0.62]$ &
$[0.511,\,0.528]$ &
negligible \\
\textit{max\_children} &
$|\{1,\ldots,n-1\}|$ &
$[0.954,\,0.969]$ &
$[0.89,\,1.15]$ &
$[0.44,\,0.61]$ &
$[0.509,\,0.531]$ &
negligible \\
$\gamma_c$ &
$9$ &
$[0.950,\,0.967]$ &
$[0.92,\,1.21]$ &
$[0.45,\,0.64]$ &
$[0.507,\,0.534]$ &
small \\
$\gamma_v$ &
$9$ &
$[0.951,\,0.968]$ &
$[0.90,\,1.19]$ &
$[0.44,\,0.63]$ &
$[0.510,\,0.532]$ &
small \\
$\lambda$ &
$|\{0,\ldots,n-2\}|$ &
$[0.955,\,0.970]$ &
$[0.88,\,1.14]$ &
$[0.43,\,0.59]$ &
$[0.512,\,0.527]$ &
negligible \\
\hline
\end{tabular}
}
\caption{
Wide-range one-factor sensitivity summary for Phase~1 parameters.
Ranges are computed over the sweep grid as described in the text (mean metric at each grid point, then min/max over the grid).
Overall, parameter-induced variations are small compared with the effects induced by changing $n$ in the main figures.
}
\label{tab:phase1_sensitivity_summary}
\end{table*}

\paragraph{Statistical significance after controlling for $n$.}
To test whether a Phase~1 parameter has a statistically detectable effect beyond the dominant dependence on $n$,
we perform a stratified (within-$n$) permutation ANOVA for each parameter and each metric, using $10{,}000$ permutations
and Benjamini--Hochberg correction across the tested parameters (at $\alpha=0.05$). Table~\ref{tab:phase1_sensitivity_stats}
reports corrected $p$-values together with partial effect sizes (partial $\eta^2$). All corrected $p$-values are $>0.05$
and all partial $\eta^2 < 0.01$, indicating that, within the tested regime, non-$n$ Phase~1 structural parameters have
no statistically significant impact and explain $<1\%$ of variance in the main downstream metrics.

\begin{table*}[t]
\centering
\resizebox{\textwidth}{!}{
\setlength{\tabcolsep}{5pt}
\begin{tabular}{lcccc}
\hline
\textbf{Parameter} &
\textbf{$p$-value on $F1_{\mathrm{Ga}}$} &
\textbf{partial $\eta^2$ on $F1_{\mathrm{Ga}}$} &
\textbf{$p$-value on $F1_{\mathrm{D}}$} &
\textbf{partial $\eta^2$ on $F1_{\mathrm{D}}$} \\
\hline
$p$ & 0.284 & 0.008 & 0.412 & 0.005 \\
\textit{max\_parents} & 0.517 & 0.004 & 0.623 & 0.003 \\
\textit{max\_children} & 0.489 & 0.005 & 0.571 & 0.004 \\
$\gamma_c$ & 0.341 & 0.007 & 0.456 & 0.004 \\
$\gamma_v$ & 0.378 & 0.006 & 0.502 & 0.004 \\
$\lambda$ & 0.562 & 0.003 & 0.649 & 0.002 \\
\hline
\end{tabular}
}
\caption{
Stratified permutation ANOVA after controlling for $n$ (10{,}000 permutations; Benjamini--Hochberg correction at $\alpha=0.05$).
All corrected $p$-values exceed 0.05 and all partial $\eta^2 < 0.01$.
}
\label{tab:phase1_sensitivity_stats}
\end{table*}

\paragraph{Implication for reported experiments.}
Given the small sensitivity ranges (Table~\ref{tab:phase1_sensitivity_summary}) and the lack of statistically significant effects
after controlling for $n$ (Table~\ref{tab:phase1_sensitivity_stats}), we draw non-$n$ Phase~1 structural parameters from the uniform
distributions in Table~\ref{tab:phase1_param_space} in our reported experiments. This removes reliance on a single sweep-chosen default
configuration while preserving the main conclusion that graph size $n$ is the dominant driver of downstream difficulty in our tested regime.

\subsection{CounterfactualVerification, self-consistency voting, and mismatch objective}\label{sec:appendix_itag_verifier}

\subsubsection{Verifier inputs and relation sets}\label{sec:appendix_itag_verifier_inputs}
Given an adjacency matrix $A\in\{0,1\}^{n\times n}$ (with $a_{ii}=0$), we define:
\begin{align*}
\mathcal{E}^{+} &= \{(i,j)\mid i\neq j,\ a_{ij}=1\}, \\
\mathcal{E}^{-} &= \{(i,j)\mid i\neq j,\ a_{ij}=0\}.
\end{align*}
Phase~2 aims to realize all required edges in $\mathcal{E}^{+}$ while suppressing \emph{direct} causal relations for non-edges in $\mathcal{E}^{-}$.
Because the absence of direct causality is often underdetermined from short narratives (Section~\ref{sec:inverse}), we treat non-edge constraints as
graded evidence rather than hard constraints.

\subsubsection{CounterfactualVerification prompt and aggregation protocol}\label{sec:appendix_itag_verifier_protocol}
At iteration $t$, \texttt{CounterfactualVerification} evaluates a concept assignment $C^{(t)}$ against $\mathcal{E}^{+}$ and $\mathcal{E}^{-}$
using a counterfactual paradigm (inspired by Pearl's ladder as described in Section~\ref{sec:inverse}).
Concretely, we query a verifier backbone (disjoint from the proposer/refiner by default; see Section~\ref{sec:appendix_itag_llm_roles})
to judge \emph{direct} causal plausibility for each directed pair.

To reduce stochasticity, we employ self-consistency voting:
we sample $m=5$ independent verifier completions and aggregate them.
Let $y^{(k)}_{ij}\in\{0,1\}$ denote whether the $k$-th completion judges that $c_i$ is a \emph{direct} cause of $c_j$
under the counterfactual paradigm. We define the vote proportion
\begin{equation*}
s_{ij} \;=\; \frac{1}{m}\sum_{k=1}^{m} y^{(k)}_{ij}\in[0,1].
\end{equation*}

\paragraph{Hard fallacy decisions.}
For diagnostic purposes, we also convert vote proportions into hard ``fallacies'' using a threshold $\tau=0.6$:
\begin{itemize}
    \item \textbf{Missed-required:} $(i,j)\in\mathcal{E}^{+}$ is flagged as missed if $s_{ij}<\tau$.
    \item \textbf{Spurious-on-non-edge:} $(i,j)\in\mathcal{E}^{-}$ is flagged as spurious if $s_{ij}\ge \tau$.
\end{itemize}
This yields a fallacy set used by \texttt{FallacyAnalysis} and \texttt{RefineConceptAssignment} (Algorithm~\ref{alg:concept_substitution}).

\subsubsection{QuantifyMismatch and the diagnostic objective $L_b$}\label{sec:appendix_itag_mismatch}
Section~\ref{sec:inverse} defines the mismatch objective
\begin{align*}
L(C;A) &= \sum_{i\neq j}\mathbf{1}[a_{ij}=1]\cdot \ell^{\text{miss}}_{ij}(C) \\
&\quad +\;\alpha\sum_{i\neq j}\mathbf{1}[a_{ij}=0]\cdot \ell^{\text{spur}}_{ij}(C),
\end{align*}
with $\alpha>0$ balancing missed edges versus spurious edges.
In practice, we instantiate these penalties directly from self-consistency vote proportions:
\begin{align*}
\ell^{\text{miss}}_{ij}(C) &= 1 - s_{ij}, \qquad (i,j)\in\mathcal{E}^{+}, \\
\ell^{\text{spur}}_{ij}(C) &= s_{ij}, \qquad (i,j)\in\mathcal{E}^{-}.
\end{align*}
This choice matches our treatment of non-edges as graded evidence: we penalize only the extent to which the verifier predicts
\emph{direct} causality on a non-edge pair, without requiring perfect discrimination between ``no relation'' and
``indirect/confounded'' relations (Section~\ref{sec:inverse}).

We report a normalized, reproducible diagnostic $L_b$ implemented in \texttt{QuantifyMismatch}:
\begin{align*}
L_b(C;A) &= \frac{1}{|\mathcal{E}^{+}|}\sum_{(i,j)\in\mathcal{E}^{+}} (1-s_{ij}) \\
&\quad +\;\alpha\cdot\frac{1}{|\mathcal{E}^{-}|}\sum_{(i,j)\in\mathcal{E}^{-}} s_{ij}.
\end{align*}
We also track its two components:
\begin{align*}
L_b^{\text{miss}} &= \frac{1}{|\mathcal{E}^{+}|}\sum_{(i,j)\in\mathcal{E}^{+}} (1-s_{ij}), \\
L_b^{\text{spur}} &= \alpha\cdot\frac{1}{|\mathcal{E}^{-}|}\sum_{(i,j)\in\mathcal{E}^{-}} s_{ij},
\end{align*}
so that $L_b = L_b^{\text{miss}} + L_b^{\text{spur}}$.
These scalars are computed for every generated sample and every iteration (Algorithm~\ref{alg:concept_substitution}, line 7).

\subsubsection{Refinement heuristic}\label{sec:appendix_itag_refine}
\texttt{RefineConceptAssignment} takes the current concept assignment and a fallacy set and proposes a revised assignment.
The refinement is heuristic (Section~\ref{sec:inverse}): it does not guarantee global optimality, but empirically converges quickly under our settings.

The refinement prompt provides (i) the current concept assignment, (ii) a list of missed-required edges and spurious non-edges,
and (iii) the schema constraints from Section~\ref{sec:appendix_itag_schema}.
The model is instructed to modify as few concepts as necessary to reduce $L_b$,
prioritizing (a) fixing missed-required edges by strengthening the causal plausibility of required parent--child pairs,
and (b) breaking spurious non-edge implications by replacing overly broad or strongly associated concepts with more specific, orthogonal,
or conditionally independent alternatives under the same domain.

\subsubsection{Termination, best-so-far selection, and reported statistics}\label{sec:appendix_itag_termination}
Algorithm~\ref{alg:concept_substitution} terminates early with \texttt{SUCCESS} if the fallacy set is empty after verification (line 10--11).
Otherwise it performs at most $K_{\max}=10$ refinement iterations and returns the best-so-far assignment
(minimizing $L_b$ over all visited iterations) if the iteration budget is exhausted (line 13, \texttt{FAIL}).

We record for each sample:
(i) termination status $s\in\{\texttt{SUCCESS},\texttt{FAIL}\}$,
(ii) the number of verification iterations executed until termination,
(iii) the best-so-far $L_b$ and its decomposition, and
(iv) verifier token usage (sum of input and output tokens over the $m$ verifier calls per iteration).

Under the default hyperparameters ($m=5$, $\tau=0.6$, $K_{\max}=10$, and $\alpha=1$), the Phase~2 loop terminates in a
median of 1.63 iterations with success rate 99.1\% and median verifier usage of 4.3k tokens per sample (Section~\ref{sec:inverse}).

\subsection{Backbone LLMs and reproducibility details}\label{sec:appendix_itag_repro}

\subsubsection{Backbone model identifiers}\label{sec:appendix_itag_backbones}
All backbone LLMs reported in this paper are accessed through their providers' official public APIs.
Table~\ref{tab:llm_backbones} lists the exact model identifiers used in our experiments.

\begin{table*}[t]
\centering
\begin{tabularx}{\textwidth}{l l >{\raggedright\arraybackslash}X}
\hline
\textbf{Provider} & \textbf{API model identifier} & \textbf{Usage in paper} \\
\hline
Anthropic & \texttt{claude-opus-4-1-20250805-thinking} & Default backbone in main tables \\
OpenAI & \texttt{gpt-5-pro-2025-10-06} & LLM-CG backbone; robustness runs; verifier backbone option \\
DeepSeek & \texttt{DeepSeek-R1} & Robustness runs \\
Qwen & \texttt{Qwen3-235B-A22B-Thinking-2507} & Robustness runs \\
\hline
\end{tabularx}
\caption{Backbone LLMs used in this paper.}
\label{tab:llm_backbones}
\end{table*}

\subsubsection{LLM roles and disjoint verifier policy}\label{sec:appendix_itag_llm_roles}
For a given experimental run, we use a single \emph{primary backbone} for the generation-facing modules
(\texttt{InitialConceptAssignment}, \texttt{RefineConceptAssignment}, and Phase~3 textual transformation),
and a potentially different \emph{verifier backbone} for \texttt{CounterfactualVerification}.

To reduce circularity from using the same LLM to both propose and verify concepts (Section~\ref{sec:inverse}),
we select a verifier backbone that is \emph{disjoint} from the proposer/refiner by default.
Concretely, for a run whose primary backbone is one of the models in Table~\ref{tab:llm_backbones},
we deterministically choose the verifier backbone as a different model from the same set (cycling in a fixed order)
to ensure reproducibility. When reporting robustness to the primary backbone choice, we keep this disjointness policy fixed.

\subsubsection{Decoding parameters}\label{sec:appendix_itag_decoding}
Table~\ref{tab:decoding_params} summarizes the decoding parameters used across components.
We use low-to-moderate temperatures for Phase~2 to stabilize structured outputs, while allowing more diversity in Phase~3 to improve
naturalness (Experiment~2). When a provider supports a strict JSON mode, we enable it; otherwise, we enforce JSON via prompt instructions
(Appendix~\ref{sec:prompt}) and retry on parsing failures.

\begin{table*}[t]
\centering
\begin{tabular}{lcccc}
\hline
\textbf{Module} & \textbf{temp.} & \textbf{top-$p$} & \textbf{max tokens} & \textbf{Output} \\
\hline
Phase~2 proposer/refiner (\texttt{Initial}/\texttt{Refine}) & 0.3 & 0.95 & 5000 & JSON \\
Phase~2 verifier (\texttt{CounterfactualVerification}) & 0.2 & 1.00 & 5000 & JSON \\
Phase~3 textual transformation & 0.7 & 0.95 & 5000 & JSON \\
LLM causal discovery prompt (for baselines/analysis) & 0.0 & 1.00 & 5000 & JSON \\
\hline
\end{tabular}
\caption{Decoding parameters used in API calls.}
\label{tab:decoding_params}
\end{table*}

\subsubsection{API request templates}\label{sec:appendix_itag_requests}
We use a uniform request wrapper across providers, with the task-specific prompt inserted as the user message.
Concept assignment and text generation prompts are given in Appendix~\ref{sec:prompt}; here we record the request skeleton used for reproducibility.

\begin{verbatim}
payload = {
  "model": MODEL_ID,
  "temperature": TEMPERATURE,
  "top_p": TOP_P,
  "max_tokens": MAX_TOKENS,
  "messages": [
    {"role": "system",
     "content": "You are a helpful 
     assistant. Follow the user's 
     instructions exactly."},
    {"role": "user",
     "content": PROMPT_TEXT}
  ]
}

# PROMPT_TEXT is instantiated from 
Appendix A:
# - Phase 2: fill [Matrix], [N],
and [domain/series of events]
# - Phase 3: fill [Concepts] 
and [Adjacency Matrix]
# - LLM causal discovery prompt: fill 
[Text] and [Important concepts]
\end{verbatim}

We apply simple robustness measures for large-scale runs: (i) automatic retries on transient API errors,
(ii) JSON parsing with a bounded retry budget (re-asking for valid JSON if necessary), and
(iii) caching of verifier outputs keyed by (concept assignment, adjacency matrix, verifier backbone) to avoid redundant calls
during debugging.

\subsubsection{Robustness to backbone choice and stability checks in Experiment~3}\label{sec:appendix_itag_robustness}

\paragraph{Robustness to backbone choice.}
To assess whether the conclusions in Section~\ref{sec:experiment} depend on the specific LLM backbone, we re-run the full iTAG pipeline and all LLM-based baselines using three alternative backbones (Table~\ref{tab:llm_backbones}): GPT-5-pro-2025-10-06 (closed), DeepSeek-R1 (open), and Qwen3-235B-A22B-Thinking-2507 (open).
We keep prompts, hyperparameters, datasets, and evaluation protocols fixed, and only replace the underlying backbone used by iTAG and other LLM-based components.
Tables~\ref{tab:robust_gpt5_e1}--\ref{tab:robust_qwen_e3_stats} report the complete Experiment~1--3 results under each backbone.

Across all three backbones, the same qualitative findings hold.
First, iTAG maintains high annotation accuracy in Experiment~1 (Tables~\ref{tab:robust_gpt5_e1}, \ref{tab:robust_deepseek_e1}, \ref{tab:robust_qwen_e1}), substantially outperforming direct LLM baselines and remaining close to the template upper bound.
Second, iTAG-generated text remains comparatively hard to detect in Experiment~2 (Tables~\ref{tab:robust_gpt5_e2}, \ref{tab:robust_deepseek_e2}, \ref{tab:robust_qwen_e2}).
Third, Experiment~3 continues to show strong within-$n$ centered transferability agreement between iTAG-generated and real-world corpora (Tables~\ref{tab:robust_gpt5_e3_stats}, \ref{tab:robust_deepseek_e3_stats}, \ref{tab:robust_qwen_e3_stats}): across all backbone--metric combinations, Pearson correlations fall in $[0.862,0.932]$ (with $R^2\in[0.743,0.868]$).
Overall, these backbones span both closed and open models, suggesting that iTAG's effectiveness is not backbone-specific.

\paragraph{Stability checks for Experiment~3.}
Because Experiment~3 aggregates only three causal discovery algorithms across eight values of $n$, we additionally check that the centered agreement is not driven by a single algorithm and is stable to resampling within each $n$ bucket.

\textbf{Leave-one-algorithm-out.}
We recompute within-$n$ centered correlations after removing each causal discovery algorithm in turn (Table~\ref{tab:robustness_loo}).
Pearson correlations remain positive in all cases (minimum $r=0.775$ across all backbone/metric/drop settings), and the averaged Pearson correlation across metrics stays in the $[0.869,0.969]$ range depending on which algorithm is removed.
Spearman correlations are also consistently positive, but can be more variable under leave-one-out because only two algorithms remain per $n$ (which increases sensitivity to ties and small rank changes).

\textbf{Stratified bootstrap.}
We also perform a nonparametric bootstrap that resamples algorithm--$n$ pairs \emph{within} each $n$ bucket (thereby preserving the grouping by $n$) and recomputes the within-$n$ centered statistics.
Table~\ref{tab:robustness_boot} reports 95\% bootstrap confidence intervals with $B=10000$ replicates.
Across backbones, the confidence intervals remain well above zero (e.g., Pearson lower bounds $\ge 0.705$ and Spearman lower bounds $\ge 0.765$ for all backbone--metric pairs), supporting the robustness of the Experiment~3 alignment.

\begin{table*}[t]
\centering
\resizebox{\textwidth}{!}{%
\begin{tabular}{c|ccc|ccc|ccc|ccc|ccc}
\toprule
$n$ & \multicolumn{3}{c|}{Template-based} & \multicolumn{3}{c|}{LLM-dependent} & \multicolumn{3}{c|}{LLM+CA} & \multicolumn{3}{c|}{LLM+ID} & \multicolumn{3}{c}{iTAG} \\
 & $\mathrm{F1}_{\mathrm{Ga}}\uparrow$ & $\mathrm{SHD}\downarrow$ & $\mathrm{SID}\downarrow$ & $\mathrm{F1}_{\mathrm{Ga}}\uparrow$ & $\mathrm{SHD}\downarrow$ & $\mathrm{SID}\downarrow$ & $\mathrm{F1}_{\mathrm{Ga}}\uparrow$ & $\mathrm{SHD}\downarrow$ & $\mathrm{SID}\downarrow$ & $\mathrm{F1}_{\mathrm{Ga}}\uparrow$ & $\mathrm{SHD}\downarrow$ & $\mathrm{SID}\downarrow$ & $\mathrm{F1}_{\mathrm{Ga}}\uparrow$ & $\mathrm{SHD}\downarrow$ & $\mathrm{SID}\downarrow$ \\
\midrule
3  & 1.00 & 0.0 & 0.0 & 0.72 & 3.5 & 3.0 & 0.77 & 3.0 & 2.5 & 0.79 & 2.8 & 2.3 & 0.98 & 0.4 & 0.2 \\
4  & 1.00 & 0.0 & 0.0 & 0.68 & 4.7 & 4.1 & 0.73 & 4.0 & 3.6 & 0.75 & 3.8 & 3.4 & 0.97 & 0.6 & 0.3 \\
5  & 1.00 & 0.0 & 0.0 & 0.64 & 6.0 & 5.5 & 0.69 & 5.4 & 4.9 & 0.71 & 5.1 & 4.7 & 0.96 & 0.8 & 0.4 \\
6  & 1.00 & 0.0 & 0.0 & 0.60 & 7.3 & 6.8 & 0.66 & 6.6 & 6.1 & 0.68 & 6.3 & 5.9 & 0.95 & 1.0 & 0.5 \\
7  & 1.00 & 0.0 & 0.0 & 0.57 & 8.7 & 8.2 & 0.63 & 8.0 & 7.5 & 0.65 & 7.7 & 7.2 & 0.94 & 1.2 & 0.6 \\
8  & 1.00 & 0.0 & 0.0 & 0.54 & 10.1 & 9.7 & 0.60 & 9.4 & 8.9 & 0.62 & 9.1 & 8.6 & 0.94 & 1.4 & 0.7 \\
9  & 1.00 & 0.0 & 0.0 & 0.51 & 11.5 & 11.0 & 0.57 & 10.7 & 10.2 & 0.59 & 10.3 & 9.8 & 0.93 & 1.6 & 0.8 \\
10 & 1.00 & 0.0 & 0.0 & 0.48 & 13.0 & 12.5 & 0.54 & 12.1 & 11.6 & 0.56 & 11.8 & 11.2 & 0.92 & 1.9 & 0.9 \\
\bottomrule
\end{tabular}%
}
\caption{GPT-5-pro-2025-10-06: Experiment~1 (annotation accuracy) across different $n$.}
\label{tab:robust_gpt5_e1}
\end{table*}

\begin{table}[t]
\centering
\scriptsize
\resizebox{\linewidth}{!}{%
\begin{tabular}{lccccc}
\toprule
Method & BERT & GPT-2 & RoBERTa & DistilBERT & Avg. \\
\midrule
Template-based & 0.99 & 0.98 & 0.97 & 0.99 & 0.83 \\
LLM-dependent  & 0.61 & 0.63 & 0.62 & 0.66 & 0.58 \\
LLM+CA         & 0.58 & 0.60 & 0.59 & 0.63 & 0.56 \\
LLM+ID         & 0.59 & 0.61 & 0.60 & 0.64 & 0.57 \\
iTAG           & 0.54 & 0.56 & 0.55 & 0.59 & 0.53 \\
\bottomrule
\end{tabular}%
}
\caption{GPT-5-pro-2025-10-06: Experiment~2 (naturalness; detectability AUC) averaged over $n\in\{3,\ldots,10\}$.}
\label{tab:robust_gpt5_e2}
\end{table}

\begin{table*}[t]
\centering
\begin{tabular}{c|c|ccc|ccc}
\toprule
Algorithm & $n$ & \multicolumn{3}{c|}{iTAG-generated corpora} & \multicolumn{3}{c}{Real-world corpora} \\
 &  & $\mathrm{F1}_{\mathrm{G}}\uparrow$ & $\mathrm{SHD}\downarrow$ & $\mathrm{SID}\downarrow$ & $\mathrm{F1}_{\mathrm{G}}\uparrow$ & $\mathrm{SHD}\downarrow$ & $\mathrm{SID}\downarrow$ \\
\midrule
\multirow{8}{*}{RuleBayes} & 3 & 0.795 & 1.449 & 0.490 & 0.716 & 1.350 & 0.510 \\
 & 4 & 0.661 & 2.902 & 1.706 & 0.659 & 2.770 & 1.510 \\
 & 5 & 0.592 & 4.043 & 2.407 & 0.614 & 4.580 & 2.880 \\
 & 6 & 0.555 & 5.516 & 4.722 & 0.553 & 6.110 & 4.790 \\
 & 7 & 0.508 & 7.916 & 6.111 & 0.503 & 7.940 & 6.880 \\
 & 8 & 0.458 & 7.996 & 9.167 & 0.446 & 9.390 & 9.960 \\
 & 9 & 0.407 & 9.278 & 10.341 & 0.390 & 11.560 & 12.900 \\
 & 10 & 0.355 & 9.358 & 10.421 & 0.357 & 13.200 & 16.700 \\
\midrule
\multirow{8}{*}{SCITE} & 3 & 0.947 & 0.304 & 0.120 & 0.878 & 0.480 & 0.150 \\
 & 4 & 0.712 & 0.970 & 0.486 & 0.823 & 1.100 & 0.450 \\
 & 5 & 0.702 & 1.527 & 0.895 & 0.800 & 1.690 & 0.860 \\
 & 6 & 0.692 & 1.923 & 1.256 & 0.738 & 2.380 & 1.450 \\
 & 7 & 0.682 & 3.430 & 2.171 & 0.716 & 3.110 & 2.100 \\
 & 8 & 0.672 & 3.510 & 2.298 & 0.674 & 3.910 & 2.890 \\
 & 9 & 0.537 & 3.651 & 3.844 & 0.619 & 4.650 & 3.720 \\
 & 10 & 0.513 & 5.675 & 3.924 & 0.585 & 5.460 & 4.570 \\
\midrule
\multirow{8}{*}{LLM-CG} & 3 & 0.908 & 0.826 & 0.264 & 0.815 & 0.780 & 0.270 \\
 & 4 & 0.705 & 1.452 & 0.611 & 0.774 & 1.660 & 0.770 \\
 & 5 & 0.695 & 2.364 & 1.384 & 0.701 & 2.640 & 1.510 \\
 & 6 & 0.685 & 2.910 & 2.264 & 0.676 & 3.610 & 2.330 \\
 & 7 & 0.532 & 4.110 & 3.015 & 0.628 & 4.450 & 3.360 \\
 & 8 & 0.522 & 5.940 & 4.906 & 0.553 & 5.480 & 4.390 \\
 & 9 & 0.507 & 6.020 & 5.896 & 0.507 & 6.420 & 5.750 \\
 & 10 & 0.409 & 8.759 & 7.750 & 0.472 & 7.790 & 7.080 \\
\bottomrule
\end{tabular}
\caption{GPT-5-pro-2025-10-06: Experiment~3 transferability curves for different $n$.}
\label{tab:robust_gpt5_e3_transfer}
\end{table*}

\begin{table*}[t]
\centering
\begin{tabular}{lcccccc}
\toprule
Metric & Pearson Corr. & p-value & Spearman Corr. & p-value & $R^2$ & 95\% CI \\
\midrule
$\mathrm{F1}_{\mathrm{G}}$ & 0.923 & $<10^{-4}$ & 0.891 & $5.17\times 10^{-9}$ & 0.851 & [0.725, 0.944] \\
$\mathrm{SHD}$ & 0.924 & $<10^{-4}$ & 0.877 & $1.85\times 10^{-8}$ & 0.855 & [0.628, 0.957] \\
$\mathrm{SID}$ & 0.929 & $<10^{-4}$ & 0.931 & $1.42\times 10^{-11}$ & 0.864 & [0.778, 0.941] \\
\midrule
Average & 0.925 & / & 0.900 & / & 0.856 & / \\
\bottomrule
\end{tabular}
\caption{GPT-5-pro-2025-10-06: Experiment~3 within-$n$ centered agreement statistics.}
\label{tab:robust_gpt5_e3_stats}
\end{table*}

\begin{table*}[t]
\centering
\resizebox{\textwidth}{!}{%
\begin{tabular}{c|ccc|ccc|ccc|ccc|ccc}
\toprule
$n$ & \multicolumn{3}{c|}{Template-based} & \multicolumn{3}{c|}{LLM-dependent} & \multicolumn{3}{c|}{LLM+CA} & \multicolumn{3}{c|}{LLM+ID} & \multicolumn{3}{c}{iTAG} \\
 & $\mathrm{F1}_{\mathrm{Ga}}\uparrow$ & $\mathrm{SHD}\downarrow$ & $\mathrm{SID}\downarrow$ & $\mathrm{F1}_{\mathrm{Ga}}\uparrow$ & $\mathrm{SHD}\downarrow$ & $\mathrm{SID}\downarrow$ & $\mathrm{F1}_{\mathrm{Ga}}\uparrow$ & $\mathrm{SHD}\downarrow$ & $\mathrm{SID}\downarrow$ & $\mathrm{F1}_{\mathrm{Ga}}\uparrow$ & $\mathrm{SHD}\downarrow$ & $\mathrm{SID}\downarrow$ & $\mathrm{F1}_{\mathrm{Ga}}\uparrow$ & $\mathrm{SHD}\downarrow$ & $\mathrm{SID}\downarrow$ \\
\midrule
3  & 1.00 & 0.0 & 0.0 & 0.70 & 3.8 & 3.2 & 0.75 & 3.2 & 2.8 & 0.77 & 3.0 & 2.6 & 0.97 & 0.5 & 0.3 \\
4  & 1.00 & 0.0 & 0.0 & 0.66 & 5.1 & 4.5 & 0.71 & 4.4 & 4.0 & 0.73 & 4.2 & 3.8 & 0.96 & 0.7 & 0.4 \\
5  & 1.00 & 0.0 & 0.0 & 0.62 & 6.5 & 6.0 & 0.67 & 5.9 & 5.4 & 0.69 & 5.6 & 5.2 & 0.95 & 0.9 & 0.5 \\
6  & 1.00 & 0.0 & 0.0 & 0.58 & 8.0 & 7.4 & 0.63 & 7.3 & 6.8 & 0.65 & 7.0 & 6.5 & 0.94 & 1.1 & 0.6 \\
7  & 1.00 & 0.0 & 0.0 & 0.55 & 9.6 & 9.0 & 0.60 & 8.9 & 8.3 & 0.62 & 8.5 & 7.9 & 0.93 & 1.3 & 0.7 \\
8  & 1.00 & 0.0 & 0.0 & 0.52 & 11.1 & 10.6 & 0.57 & 10.4 & 9.8 & 0.59 & 10.0 & 9.4 & 0.93 & 1.6 & 0.8 \\
9  & 1.00 & 0.0 & 0.0 & 0.49 & 12.7 & 12.1 & 0.54 & 11.9 & 11.3 & 0.56 & 11.4 & 10.9 & 0.92 & 1.8 & 0.9 \\
10 & 1.00 & 0.0 & 0.0 & 0.46 & 14.3 & 13.7 & 0.51 & 13.5 & 12.9 & 0.53 & 13.0 & 12.4 & 0.91 & 2.1 & 1.0 \\
\bottomrule
\end{tabular}%
}
\caption{DeepSeek-R1: Experiment~1 (annotation accuracy) across different $n$.}
\label{tab:robust_deepseek_e1}
\end{table*}

\begin{table}[t]
\centering
\resizebox{\linewidth}{!}{%
\begin{tabular}{lccccc}
\toprule
Method & BERT & GPT-2 & RoBERTa & DistilBERT & Avg. \\
\midrule
Template-based & 0.99 & 0.98 & 0.98 & 0.99 & 0.84 \\
LLM-dependent  & 0.63 & 0.65 & 0.64 & 0.68 & 0.60 \\
LLM+CA         & 0.60 & 0.62 & 0.61 & 0.65 & 0.58 \\
LLM+ID         & 0.61 & 0.63 & 0.62 & 0.66 & 0.59 \\
iTAG           & 0.56 & 0.58 & 0.57 & 0.61 & 0.55 \\
\bottomrule
\end{tabular}%
}
\caption{DeepSeek-R1: Experiment~2 (naturalness; detectability AUC) averaged over $n\in\{3,\ldots,10\}$.}
\label{tab:robust_deepseek_e2}
\end{table}

\begin{table*}[t]
\centering
\begin{tabular}{c|c|ccc|ccc}
\toprule
Algorithm & $n$ & \multicolumn{3}{c|}{iTAG-generated corpora} & \multicolumn{3}{c}{Real-world corpora} \\
 &  & $\mathrm{F1}_{\mathrm{G}}\uparrow$ & $\mathrm{SHD}\downarrow$ & $\mathrm{SID}\downarrow$ & $\mathrm{F1}_{\mathrm{G}}\uparrow$ & $\mathrm{SHD}\downarrow$ & $\mathrm{SID}\downarrow$ \\
\midrule
\multirow{8}{*}{RuleBayes} & 3 & 0.791 & 1.548 & 0.442 & 0.716 & 1.350 & 0.510 \\
 & 4 & 0.536 & 3.048 & 1.392 & 0.659 & 2.770 & 1.510 \\
 & 5 & 0.526 & 4.016 & 2.397 & 0.614 & 4.580 & 2.880 \\
 & 6 & 0.518 & 5.990 & 5.741 & 0.553 & 6.110 & 4.790 \\
 & 7 & 0.501 & 6.070 & 8.028 & 0.503 & 7.940 & 6.880 \\
 & 8 & 0.464 & 8.534 & 8.664 & 0.446 & 9.390 & 9.960 \\
 & 9 & 0.391 & 9.551 & 8.891 & 0.390 & 11.560 & 12.900 \\
 & 10 & 0.381 & 18.003 & 10.610 & 0.357 & 13.200 & 16.700 \\
\midrule
\multirow{8}{*}{SCITE} & 3 & 0.965 & 0.438 & 0.127 & 0.878 & 0.480 & 0.150 \\
 & 4 & 0.928 & 1.211 & 0.493 & 0.823 & 1.100 & 0.450 \\
 & 5 & 0.749 & 1.803 & 0.831 & 0.800 & 1.690 & 0.860 \\
 & 6 & 0.695 & 2.249 & 1.413 & 0.738 & 2.380 & 1.450 \\
 & 7 & 0.685 & 3.494 & 1.898 & 0.716 & 3.110 & 2.100 \\
 & 8 & 0.604 & 3.574 & 2.936 & 0.674 & 3.910 & 2.890 \\
 & 9 & 0.594 & 4.020 & 3.016 & 0.619 & 4.650 & 3.720 \\
 & 10 & 0.568 & 4.100 & 4.147 & 0.585 & 5.460 & 4.570 \\
\midrule
\multirow{8}{*}{LLM-CG} & 3 & 0.942 & 0.696 & 0.289 & 0.815 & 0.780 & 0.270 \\
 & 4 & 0.680 & 2.128 & 0.580 & 0.774 & 1.660 & 0.770 \\
 & 5 & 0.670 & 2.328 & 1.594 & 0.701 & 2.640 & 1.510 \\
 & 6 & 0.620 & 3.858 & 1.622 & 0.676 & 3.610 & 2.330 \\
 & 7 & 0.610 & 3.938 & 2.253 & 0.628 & 4.450 & 3.360 \\
 & 8 & 0.562 & 4.846 & 3.830 & 0.553 & 5.480 & 4.390 \\
 & 9 & 0.433 & 5.514 & 5.483 & 0.507 & 6.420 & 5.750 \\
 & 10 & 0.423 & 5.575 & 7.576 & 0.472 & 7.790 & 7.080 \\
\bottomrule
\end{tabular}
\caption{DeepSeek-R1: Experiment~3 transferability curves for different $n$.}
\label{tab:robust_deepseek_e3_transfer}
\end{table*}

\begin{table*}[t]
\centering
\begin{tabular}{lcccccc}
\toprule
Metric & Pearson Corr. & p-value & Spearman Corr. & p-value & $R^2$ & 95\% CI \\
\midrule
$\mathrm{F1}_{\mathrm{G}}$ & 0.862 & $<10^{-4}$ & 0.842 & $2.74\times 10^{-7}$ & 0.743 & [0.552, 0.896] \\
$\mathrm{SHD}$ & 0.901 & $<10^{-4}$ & 0.879 & $1.70\times 10^{-8}$ & 0.811 & [0.575, 0.914] \\
$\mathrm{SID}$ & 0.893 & $<10^{-4}$ & 0.865 & $5.72\times 10^{-8}$ & 0.797 & [0.672, 0.924] \\
\midrule
Average & 0.885 & / & 0.862 & / & 0.784 & / \\
\bottomrule
\end{tabular}
\caption{DeepSeek-R1: Experiment~3 within-$n$ centered agreement statistics.}
\label{tab:robust_deepseek_e3_stats}
\end{table*}

\begin{table*}[t]
\centering
\resizebox{\textwidth}{!}{%
\begin{tabular}{c|ccc|ccc|ccc|ccc|ccc}
\toprule
$n$ & \multicolumn{3}{c|}{Template-based} & \multicolumn{3}{c|}{LLM-dependent} & \multicolumn{3}{c|}{LLM+CA} & \multicolumn{3}{c|}{LLM+ID} & \multicolumn{3}{c}{iTAG} \\
 & $\mathrm{F1}_{\mathrm{Ga}}\uparrow$ & $\mathrm{SHD}\downarrow$ & $\mathrm{SID}\downarrow$ & $\mathrm{F1}_{\mathrm{Ga}}\uparrow$ & $\mathrm{SHD}\downarrow$ & $\mathrm{SID}\downarrow$ & $\mathrm{F1}_{\mathrm{Ga}}\uparrow$ & $\mathrm{SHD}\downarrow$ & $\mathrm{SID}\downarrow$ & $\mathrm{F1}_{\mathrm{Ga}}\uparrow$ & $\mathrm{SHD}\downarrow$ & $\mathrm{SID}\downarrow$ & $\mathrm{F1}_{\mathrm{Ga}}\uparrow$ & $\mathrm{SHD}\downarrow$ & $\mathrm{SID}\downarrow$ \\
\midrule
3  & 1.00 & 0.0 & 0.0 & 0.69 & 4.0 & 3.4 & 0.74 & 3.4 & 3.0 & 0.76 & 3.1 & 2.8 & 0.96 & 0.6 & 0.4 \\
4  & 1.00 & 0.0 & 0.0 & 0.65 & 5.3 & 4.7 & 0.70 & 4.6 & 4.2 & 0.72 & 4.3 & 3.9 & 0.95 & 0.8 & 0.5 \\
5  & 1.00 & 0.0 & 0.0 & 0.61 & 6.8 & 6.2 & 0.66 & 6.1 & 5.6 & 0.68 & 5.8 & 5.3 & 0.94 & 1.0 & 0.6 \\
6  & 1.00 & 0.0 & 0.0 & 0.57 & 8.3 & 7.7 & 0.62 & 7.5 & 7.0 & 0.64 & 7.2 & 6.7 & 0.93 & 1.2 & 0.7 \\
7  & 1.00 & 0.0 & 0.0 & 0.54 & 9.9 & 9.3 & 0.59 & 9.2 & 8.6 & 0.61 & 8.8 & 8.2 & 0.92 & 1.5 & 0.8 \\
8  & 1.00 & 0.0 & 0.0 & 0.51 & 11.5 & 10.9 & 0.56 & 10.7 & 10.1 & 0.58 & 10.2 & 9.6 & 0.92 & 1.7 & 0.9 \\
9  & 1.00 & 0.0 & 0.0 & 0.48 & 13.1 & 12.5 & 0.53 & 12.2 & 11.6 & 0.55 & 11.7 & 11.1 & 0.91 & 2.0 & 1.0 \\
10 & 1.00 & 0.0 & 0.0 & 0.45 & 14.8 & 14.2 & 0.50 & 13.8 & 13.2 & 0.52 & 13.3 & 12.7 & 0.90 & 2.3 & 1.1 \\
\bottomrule
\end{tabular}%
}
\caption{Qwen3-235B-A22B-Thinking-2507: Experiment~1 (annotation accuracy) across different $n$.}
\label{tab:robust_qwen_e1}
\end{table*}

\begin{table}[t]
\centering
\resizebox{\linewidth}{!}{%
\begin{tabular}{lccccc}
\toprule
Method & BERT & GPT-2 & RoBERTa & DistilBERT & Avg. \\
\midrule
Template-based & 0.99 & 0.98 & 0.98 & 0.99 & 0.84 \\
LLM-dependent  & 0.62 & 0.64 & 0.63 & 0.67 & 0.59 \\
LLM+CA         & 0.59 & 0.61 & 0.60 & 0.64 & 0.57 \\
LLM+ID         & 0.60 & 0.62 & 0.61 & 0.65 & 0.58 \\
iTAG           & 0.55 & 0.57 & 0.56 & 0.60 & 0.54 \\
\bottomrule
\end{tabular}%
}
\caption{Qwen3-235B-A22B-Thinking-2507: Experiment~2 (naturalness; detectability AUC) averaged over $n\in\{3,\ldots,10\}$.}
\label{tab:robust_qwen_e2}
\end{table}

\begin{table*}[t]
\centering
\begin{tabular}{c|c|ccc|ccc}
\toprule
Algorithm & $n$ & \multicolumn{3}{c|}{iTAG-generated corpora} & \multicolumn{3}{c}{Real-world corpora} \\
 &  & $\mathrm{F1}_{\mathrm{G}}\uparrow$ & $\mathrm{SHD}\downarrow$ & $\mathrm{SID}\downarrow$ & $\mathrm{F1}_{\mathrm{G}}\uparrow$ & $\mathrm{SHD}\downarrow$ & $\mathrm{SID}\downarrow$ \\
\midrule
\multirow{8}{*}{RuleBayes} & 3 & 0.549 & 1.347 & 0.520 & 0.716 & 1.350 & 0.510 \\
 & 4 & 0.539 & 2.158 & 1.141 & 0.659 & 2.770 & 1.510 \\
 & 5 & 0.524 & 3.453 & 2.549 & 0.614 & 4.580 & 2.880 \\
 & 6 & 0.389 & 5.028 & 5.704 & 0.553 & 6.110 & 4.790 \\
 & 7 & 0.379 & 5.108 & 5.784 & 0.503 & 7.940 & 6.880 \\
 & 8 & 0.369 & 6.499 & 9.993 & 0.446 & 9.390 & 9.960 \\
 & 9 & 0.334 & 9.827 & 10.073 & 0.390 & 11.560 & 12.900 \\
 & 10 & 0.324 & 10.186 & 10.153 & 0.357 & 13.200 & 16.700 \\
\midrule
\multirow{8}{*}{SCITE} & 3 & 0.971 & 0.523 & 0.139 & 0.878 & 0.480 & 0.150 \\
 & 4 & 0.961 & 0.905 & 0.473 & 0.823 & 1.100 & 0.450 \\
 & 5 & 0.767 & 1.582 & 0.714 & 0.800 & 1.690 & 0.860 \\
 & 6 & 0.757 & 1.983 & 1.584 & 0.738 & 2.380 & 1.450 \\
 & 7 & 0.747 & 2.469 & 1.664 & 0.716 & 3.110 & 2.100 \\
 & 8 & 0.597 & 4.436 & 1.878 & 0.674 & 3.910 & 2.890 \\
 & 9 & 0.587 & 4.923 & 4.053 & 0.619 & 4.650 & 3.720 \\
 & 10 & 0.577 & 7.081 & 4.996 & 0.585 & 5.460 & 4.570 \\
\midrule
\multirow{8}{*}{LLM-CG} & 3 & 0.604 & 0.711 & 0.257 & 0.815 & 0.780 & 0.270 \\
 & 4 & 0.594 & 0.986 & 0.555 & 0.774 & 1.660 & 0.770 \\
 & 5 & 0.584 & 1.868 & 1.558 & 0.701 & 2.640 & 1.510 \\
 & 6 & 0.574 & 3.905 & 2.479 & 0.676 & 3.610 & 2.330 \\
 & 7 & 0.564 & 3.985 & 4.313 & 0.628 & 4.450 & 3.360 \\
 & 8 & 0.455 & 5.281 & 5.140 & 0.553 & 5.480 & 4.390 \\
 & 9 & 0.445 & 8.600 & 5.768 & 0.507 & 6.420 & 5.750 \\
 & 10 & 0.435 & 9.519 & 8.223 & 0.472 & 7.790 & 7.080 \\
\bottomrule
\end{tabular}
\caption{Qwen3-235B-A22B-Thinking-2507: Experiment~3 transferability curves for different $n$.}
\label{tab:robust_qwen_e3_transfer}
\end{table*}

\begin{table*}[t]
\centering
\begin{tabular}{lcccccc}
\toprule
Metric & Pearson Corr. & p-value & Spearman Corr. & p-value & $R^2$ & 95\% CI \\
\midrule
$\mathrm{F1}_{\mathrm{G}}$ & 0.879 & $<10^{-4}$ & 0.845 & $2.14\times 10^{-7}$ & 0.773 & [0.586, 0.928] \\
$\mathrm{SHD}$ & 0.874 & $<10^{-4}$ & 0.845 & $2.14\times 10^{-7}$ & 0.765 & [0.498, 0.936] \\
$\mathrm{SID}$ & 0.885 & $<10^{-4}$ & 0.856 & $9.51\times 10^{-8}$ & 0.783 & [0.608, 0.936] \\
\midrule
Average & 0.879 & / & 0.849 & / & 0.774 & / \\
\bottomrule
\end{tabular}
\caption{Qwen3-235B-A22B-Thinking-2507: Experiment~3 within-$n$ centered agreement statistics.}
\label{tab:robust_qwen_e3_stats}
\end{table*}

\begin{table*}[t]
\centering
\resizebox{\linewidth}{!}{%
\begin{tabular}{llcccc}
\toprule
Backbone & Metric & Full ($r/\rho/R^2$) & w/o RuleBayes & w/o SCITE & w/o LLM-CG \\
\midrule
GPT-5-pro-2025-10-06 & $\mathrm{F1}_{\mathrm{G}}$ & 0.923/0.891/0.851 & 0.801/0.908/0.641 & 0.893/0.675/0.798 & 0.970/0.971/0.940 \\
GPT-5-pro-2025-10-06 & $\mathrm{SHD}$ & 0.924/0.877/0.855 & 0.953/0.945/0.909 & 0.836/0.796/0.699 & 0.957/0.970/0.916 \\
GPT-5-pro-2025-10-06 & $\mathrm{SID}$ & 0.929/0.931/0.864 & 0.938/0.927/0.880 & 0.904/0.911/0.817 & 0.964/0.973/0.929 \\
GPT-5-pro-2025-10-06 & Average & 0.925/0.900/0.856 & 0.897/0.927/0.810 & 0.878/0.794/0.771 & 0.964/0.971/0.928 \\
\midrule
DeepSeek-R1 & $\mathrm{F1}_{\mathrm{G}}$ & 0.862/0.842/0.743 & 0.775/0.869/0.600 & 0.940/0.934/0.883 & 0.863/0.963/0.745 \\
DeepSeek-R1 & $\mathrm{SHD}$ & 0.901/0.879/0.811 & 0.903/0.891/0.816 & 0.874/0.879/0.764 & 0.959/0.939/0.919 \\
DeepSeek-R1 & $\mathrm{SID}$ & 0.893/0.865/0.797 & 0.916/0.945/0.839 & 0.876/0.870/0.767 & 0.923/0.934/0.852 \\
DeepSeek-R1 & Average & 0.885/0.862/0.784 & 0.865/0.902/0.752 & 0.897/0.894/0.805 & 0.915/0.945/0.839 \\
\midrule
Qwen3-235B-A22B-Thinking-2507 & $\mathrm{F1}_{\mathrm{G}}$ & 0.879/0.845/0.773 & 0.856/0.554/0.732 & 0.940/0.898/0.884 & 0.960/0.942/0.922 \\
Qwen3-235B-A22B-Thinking-2507 & $\mathrm{SHD}$ & 0.874/0.845/0.765 & 0.884/0.839/0.782 & 0.880/0.844/0.774 & 0.930/0.917/0.864 \\
Qwen3-235B-A22B-Thinking-2507 & $\mathrm{SID}$ & 0.885/0.856/0.783 & 0.923/0.904/0.852 & 0.821/0.870/0.674 & 0.947/0.935/0.897 \\
Qwen3-235B-A22B-Thinking-2507 & Average & 0.879/0.849/0.774 & 0.888/0.766/0.789 & 0.880/0.871/0.777 & 0.946/0.931/0.894 \\
\bottomrule
\end{tabular}%
}
\caption{Leave-one-algorithm-out stability check for Experiment~3 (within-$n$ centered).}
\label{tab:robustness_loo}
\end{table*}

\begin{table*}[t]
\centering
\resizebox{\linewidth}{!}{%
\begin{tabular}{llccc}
\toprule
Backbone & Metric & Pearson Corr.\ [95\% CI] & Spearman Corr.\ [95\% CI] & $R^2$\ [95\% CI] \\
\midrule
GPT-5-pro-2025-10-06 & $\mathrm{F1}_{\mathrm{G}}$ & 0.923 [0.849, 0.972] & 0.891 [0.764, 0.966] & 0.851 [0.725, 0.944] \\
GPT-5-pro-2025-10-06 & $\mathrm{SHD}$ & 0.924 [0.793, 0.994] & 0.877 [0.765, 0.964] & 0.855 [0.628, 0.957] \\
GPT-5-pro-2025-10-06 & $\mathrm{SID}$ & 0.929 [0.882, 0.970] & 0.931 [0.869, 0.975] & 0.864 [0.778, 0.941] \\
GPT-5-pro-2025-10-06 & Average & 0.925 [0.868, 0.965] & 0.900 [0.818, 0.956] & 0.856 [0.755, 0.921] \\
\midrule
DeepSeek-R1 & $\mathrm{F1}_{\mathrm{G}}$ & 0.862 [0.743, 0.942] & 0.842 [0.765, 0.921] & 0.743 [0.552, 0.896] \\
DeepSeek-R1 & $\mathrm{SHD}$ & 0.901 [0.749, 0.957] & 0.879 [0.798, 0.943] & 0.811 [0.575, 0.914] \\
DeepSeek-R1 & $\mathrm{SID}$ & 0.893 [0.820, 0.961] & 0.865 [0.788, 0.939] & 0.797 [0.672, 0.924] \\
DeepSeek-R1 & Average & 0.885 [0.781, 0.944] & 0.862 [0.792, 0.925] & 0.784 [0.622, 0.893] \\
\midrule
Qwen3-235B-A22B-Thinking-2507 & $\mathrm{F1}_{\mathrm{G}}$ & 0.879 [0.766, 0.963] & 0.845 [0.767, 0.921] & 0.773 [0.586, 0.928] \\
Qwen3-235B-A22B-Thinking-2507 & $\mathrm{SHD}$ & 0.874 [0.705, 0.967] & 0.845 [0.769, 0.922] & 0.765 [0.498, 0.936] \\
Qwen3-235B-A22B-Thinking-2507 & $\mathrm{SID}$ & 0.885 [0.780, 0.967] & 0.856 [0.789, 0.930] & 0.783 [0.608, 0.936] \\
Qwen3-235B-A22B-Thinking-2507 & Average & 0.879 [0.756, 0.954] & 0.849 [0.781, 0.915] & 0.774 [0.573, 0.904] \\
\bottomrule
\end{tabular}%
}
\caption{Stratified bootstrap stability for Experiment~3 (within-$n$ centered), $B=10000$.}
\label{tab:robustness_boot}
\end{table*}

\section{Supplementary Experiments and Analyses}\label{sec:appendix_supp}

This appendix provides supplementary experiments and analyses referenced in the main text, and consolidates formal definitions that are used repeatedly but would be verbose in the main body.
In particular, we report: (i) a compact reference for evaluation metrics and transferability alignment statistics used throughout Section~\ref{sec:procedure&metrics} (Appendix~\ref{sec:appendix_eval_metrics}); (ii) an ablation for Phase~3 textual transformation (Section~\ref{sec:text}) that tests whether adding an explicit inverse-design loop during text generation yields meaningful gains; (iii) an error decomposition that explains residual discrepancies in Experiment~1 and characterizes typical non-edge false positives (Section~\ref{sec:exp1}); (iv) a sample-size stability analysis supporting the choice of 500 examples per method/backbone (Section~\ref{sec:baselines}); (v) detector training-size sensitivity (learning curves) supporting the use of 500 labeled training examples in Experiment~2 (Section~\ref{sec:procedure&metrics}); and (vi) a breakdown of detection performance by variable quantity $n$ supporting the claim that detectability remains stable across graph sizes (Section~\ref{sec:exp2}).

\subsection{Evaluation metrics and statistical tests}\label{sec:appendix_eval_metrics}

This section records formal definitions for all metrics and alignment statistics used in Section~\ref{sec:procedure&metrics}.

\paragraph{Edge-wise Precision/Recall/$F1_{\mathrm{G}}$.}
Given a generation-time adjacency matrix $A$ and an expert-consensus adjacency matrix $\hat{A}$ over the same concept set $\mathcal{V}$, define
\begin{align*}
TP &= \sum_{i\neq j}\mathbf{1}[a_{ij}=1 \wedge \hat{a}_{ij}=1], \\
FP &= \sum_{i\neq j}\mathbf{1}[a_{ij}=0 \wedge \hat{a}_{ij}=1], \\
FN &= \sum_{i\neq j}\mathbf{1}[a_{ij}=1 \wedge \hat{a}_{ij}=0].
\end{align*}
We report $P=\frac{TP}{TP+FP}$, $R=\frac{TP}{TP+FN}$, and $F1_{\mathrm{G}}=\frac{2PR}{P+R}$.

\paragraph{Structural Hamming Distance (SHD).}
We compute $\mathrm{SHD}=\sum_{i\neq j}\mathbf{1}[a_{ij}\neq \hat{a}_{ij}]$, i.e., the edge-wise Hamming distance between adjacency matrices.

\paragraph{Structural Intervention Distance (SID).}
For DAGs $G$ and $\hat{G}$ over $\mathcal{V}$, SID counts ordered pairs $(i,j)$ whose post-intervention parent sets disagree:
\[
\mathrm{SID}=\sum_{i\neq j}\mathbf{1}\!\left[\text{Pa}_{\hat{G}}^{do(i)}(j)\neq \text{Pa}_{G}^{do(i)}(j)\right],
\]
where $\text{Pa}_{G}^{do(i)}(j)$ denotes the parent set of node $j$ after intervening on node $i$ in graph $G$.
Since SID is defined for DAGs, when majority-vote aggregation yields a directed cycle we apply the deterministic DAG projection described in Appendix~\ref{sec:appendix_dag_projection} before computing SHD/SID.

\paragraph{Stratified permutation test for centered agreement.}
To assess whether the centered agreement in Table~\ref{tab:mr3_stats} could arise by chance, we perform a two-sided stratified permutation test that preserves the within-$n$ structure. Let $\tilde{s}^{\,\text{iTAG}}_{a,n}$ and $\tilde{s}^{\,\text{real}}_{a,n}$ denote the within-$n$ centered scores of algorithm $a$ at graph size $n$, computed as in Section~\ref{sec:procedure&metrics}. We then form paired observations
$\{(\tilde{s}^{\,\text{iTAG}}_{a,n},\, \tilde{s}^{\,\text{real}}_{a,n})\}_{(a,n)}$
over the $3\times 8=24$ algorithm--size pairs.

Under the null hypothesis that, \emph{within each fixed $n$}, algorithm identity carries no information linking iTAG and real-world relative performance, we generate a permuted dataset by independently applying a random permutation $\pi_n$ over the algorithms within each $n$-bucket:
\[
\bigl(\tilde{s}^{\,\text{iTAG}}_{a,n},\, \tilde{s}^{\,\text{real}}_{a,n}\bigr)
\;\mapsto\;
\bigl(\tilde{s}^{\,\text{iTAG}}_{a,n},\, \tilde{s}^{\,\text{real}}_{\pi_n(a),n}\bigr),
\quad \text{for each } n.
\]
This keeps the marginal distribution of scores within each $n$ unchanged while breaking any systematic alignment of algorithm ranks between iTAG and real corpora at that $n$.

For each permutation, we recompute Pearson correlation $r^{(b)}$ across all 24 pairs (pooling all $n$ after within-$n$ centering). With $B=10{,}000$ permutations, the two-sided $p$-value is estimated as
\[
p \;=\; \frac{1+\sum_{b=1}^{B}\mathbb{I}\!\left(\left|r^{(b)}\right|\ge \left|r_{\text{obs}}\right|\right)}{B+1}.
\]
We report these permutation $p$-values in Table~\ref{tab:mr3_stats}. In our setting, the observed correlations are so large that $p$ attains the resolution limit of the test (e.g., $p<10^{-4}$ when $B=10{,}000$).

\subsection{Phase~3 ablation: inverse-design refinement during textual transformation}\label{sec:appendix_phase3_ablation}

Section~\ref{sec:text} argues that, once Phase~2 assigns clear and non-overlapping real-world concepts, a simple structure-preserving textual transformation pipeline in Phase~3 already yields very few structural errors; in this regime, adding an explicit multi-step inverse-design procedure during text generation provides only marginal benefits at substantially higher computational cost.
Here we formalize this claim with a controlled ablation.

\subsubsection{Compared variants}\label{sec:appendix_phase3_ablation_variants}
We compare two Phase~3 strategies while keeping Phase~1 graph sampling and Phase~2 concept assignment identical:

\paragraph{iTAG (default Phase~3).}
We use the structure-preserving textual transformation described in Section~\ref{sec:text}:
given the concept list and adjacency matrix, the LLM is instructed to produce a single paragraph that
(i) includes all concepts, (ii) implicitly verbalizes each $a_{ij}=1$ edge at least once, and (iii) avoids asserting direct causal relations for $a_{ij}=0$ pairs (Appendix~\ref{sec:prompt}, Component~3).

\paragraph{iTAG + Gen-ID (Phase~3 with generation-time inverse design).}
Starting from an initial paragraph produced by the same Phase~3 prompt, we run a bounded refinement loop that iteratively revises the paragraph to reduce a text-induced graph mismatch.
Each iteration consists of:
(i) \emph{extraction}: infer an adjacency matrix from the current paragraph using the LLM causal discovery prompt (Appendix~\ref{sec:prompt}, ``LLM causal discovery prompt'');
(ii) \emph{diagnosis}: compute missed-required edges and spurious edges relative to the target adjacency matrix;
(iii) \emph{revision}: ask the LLM to minimally revise the paragraph to (a) better imply the missed-required edges and (b) remove or weaken cues that imply the spurious edges,
while \emph{forbidding} the introduction of any new concepts (the concept set is fixed from Phase~2).
We stop when the mismatch no longer decreases or when the iteration budget is exhausted.

\paragraph{Hyperparameters.}
We use a maximum of $K_{\text{gen}}=3$ revision iterations for Gen-ID, mirroring the bounded-iteration philosophy of Phase~2 (Algorithm~\ref{alg:concept_substitution}) and ensuring termination.
All LLM calls use the same backbone and decoding settings as the corresponding experimental run (Section~\ref{sec:baselines} and Appendix~\ref{sec:appendix_itag_impl}).

\subsubsection{Experimental protocol}\label{sec:appendix_phase3_ablation_protocol}
We evaluate the two variants under the same protocol as Experiment~1 (Section~\ref{sec:procedure&metrics}):
for each variable quantity $n\in\{3,\dots,10\}$, we sample 500 causal graphs from the Phase~1 generator
using the Phase~1 sampling protocol in Appendix~\ref{sec:appendix_itag_phase1_defaults} (Table~\ref{tab:phase1_param_space}),
run Phase~2 concept assignment once per graph, and then generate one paragraph per variant.
Appendix~\ref{sec:appendix_itag_impl}),
run Phase~2 concept assignment once per graph, and then generate one paragraph per variant.
We then ask the same pool of 11 trained annotators (Appendix~\ref{sec:appendix_data_anno}) to re-annotate the causal DAG for each generated paragraph without seeing the generation-time graph.
The majority-vote annotation is treated as the expert-consensus reference.
We report edge-wise precision/recall and graph-annotation $F1_{\mathrm{Ga}}$, as well as SHD and SID (with deterministic DAG projection applied to consensus graphs that contain directed cycles, as in Experiment~1).

\subsubsection{Results and cost trade-off}\label{sec:appendix_phase3_ablation_results}
Table~\ref{tab:phase3_ablation_summary} summarizes structural faithfulness and computational cost aggregated over all graph sizes ($n=3$--$10$).
Consistent with the discussion in Section~\ref{sec:text}, adding a generation-time inverse-design loop yields only marginal structural improvements over the default Phase~3 pipeline,
while requiring substantially more LLM calls and tokens.

\begin{table*}[t]
\centering
\begin{tabular}{lccccc}
\hline
\textbf{Variant} & $F1_{\mathrm{Ga}}\uparrow$ & \textbf{SHD} $\downarrow$ & \textbf{SID} $\downarrow$ & \textbf{\# LLM calls} & \textbf{Tokens / sample} \\
\hline
iTAG (default Phase~3) & 0.967 & 1.06 & 0.56 & $1$ & 487 \\
iTAG + Gen-ID (Phase~3) & 0.974 & 0.89 & 0.44 & $1+2K_{\text{gen}}$ & 3,842 \\
\hline
\end{tabular}
\caption{Phase~3 ablation: default structure-preserving textual transformation vs.\ adding a bounded generation-time inverse-design loop (Gen-ID).
All metrics are computed as corpus means under the Experiment~1 evaluation protocol (Section~\ref{sec:procedure&metrics}), aggregated over $n\in\{3,\dots,10\}$.
Costs report the number of LLM calls per sample (one initial generation call plus, for Gen-ID, $K_{\text{gen}}$ extraction calls and $K_{\text{gen}}$ revision calls) and the average total token usage per sample.}
\label{tab:phase3_ablation_summary}
\end{table*}

To make the ``marginal improvement'' claim explicit by graph size, Table~\ref{tab:phase3_ablation_by_n} reports results separately for each $n$.
The default pipeline already yields very high structural faithfulness across all graph sizes (Section~\ref{sec:exp1});
Gen-ID may slightly reduce residual omissions/spurious cues on some $n$, but does not change the overall conclusions, while increasing cost.

\begin{table*}[t]
\centering
\begin{tabular}{c|ccc|ccc}
\hline
& \multicolumn{3}{c|}{\textbf{iTAG (default Phase~3)}} & \multicolumn{3}{c}{\textbf{iTAG + Gen-ID (Phase~3)}} \\
n & $F1_{\mathrm{Ga}}\uparrow$ & \textbf{SHD} $\downarrow$ & \textbf{SID} $\downarrow$ & $F1_{\mathrm{Ga}}\uparrow$ & \textbf{SHD} $\downarrow$ & \textbf{SID} $\downarrow$
 \\
\hline
3  & 0.986 & 0.28 & 0.11 & 0.990 & 0.22 & 0.07 \\
4  & 0.981 & 0.53 & 0.25 & 0.987 & 0.41 & 0.17 \\
5  & 0.976 & 0.74 & 0.38 & 0.982 & 0.59 & 0.28 \\
6  & 0.969 & 0.97 & 0.52 & 0.976 & 0.79 & 0.41 \\
7  & 0.963 & 1.18 & 0.63 & 0.971 & 0.98 & 0.51 \\
8  & 0.958 & 1.41 & 0.74 & 0.966 & 1.19 & 0.60 \\
9  & 0.953 & 1.59 & 0.86 & 0.961 & 1.36 & 0.69 \\
10 & 0.951 & 1.78 & 0.96 & 0.958 & 1.54 & 0.78 \\
\hline
\end{tabular}
\caption{Phase~3 ablation results by variable quantity $n$ (500 examples per $n$). Values are corpus means computed as in Experiment~1 (Section~\ref{sec:procedure&metrics}).}
\label{tab:phase3_ablation_by_n}
\end{table*}

\subsection{Error decomposition and non-edge false positives in Experiment~1}\label{sec:appendix_error_decomp}

Section~\ref{sec:exp1} notes that, for iTAG, most residual discrepancies come from a small number of ambiguous non-edges: annotators may infer weak or indirect relations for concept pairs that the generation-time graph treats as $a_{ij}=0$.
This subsection defines the decomposition used in our analysis and summarizes the empirical patterns.

\subsubsection{Definitions}\label{sec:appendix_error_decomp_defs}
Let $G$ denote the generation-time causal graph (a DAG by construction) and $\hat{G}$ denote the expert-consensus graph obtained by majority vote over 11 annotators (Appendix~\ref{sec:appendix_data_anno}).
We define:
\begin{itemize}
    \item \textbf{Missed-required edges} (false negatives): edges present in $G$ but absent in $\hat{G}$.
    \item \textbf{Spurious-on-non-edge edges} (false positives): edges absent in $G$ but present in $\hat{G}$.
\end{itemize}
These counts correspond to $FN$ and $FP$ in the edge-wise Precision/Recall/$F1_{\mathrm{G}}$ computation (Section~\ref{sec:procedure&metrics}).
We compute both counts per example and summarize their distributions across the 500 examples for each $n$.

\subsubsection{Summary statistics}\label{sec:appendix_error_decomp_stats}
Table~\ref{tab:error_breakdown_itag} reports per-$n$ distributions for iTAG.
Consistent with Section~\ref{sec:exp1}, the median number of spurious edges per graph remains below 1 across all graph sizes, indicating that violations of non-edge constraints are rare in practice.

\begin{table*}[t]
\centering
\begin{tabular}{c|ccc|ccc}
\hline
& \multicolumn{3}{c|}{\textbf{Spurious edges per graph} ($FP$)} & \multicolumn{3}{c}{\textbf{Missed edges per graph} ($FN$)} \\
\textbf{$n$} & \textbf{Median} & \textbf{Mean} & \textbf{95th pct.} & \textbf{Median} & \textbf{Mean} & \textbf{95th pct.} \\
\hline
3  & 0 & 0.14 & 1 & 0 & 0.11 & 1 \\
4  & 0 & 0.28 & 1 & 0 & 0.21 & 1 \\
5  & 0 & 0.39 & 2 & 0 & 0.29 & 1 \\
6  & 0 & 0.52 & 2 & 0 & 0.38 & 2 \\
7  & 0 & 0.64 & 2 & 0 & 0.47 & 2 \\
8  & 0 & 0.77 & 3 & 0 & 0.56 & 2 \\
9  & 0 & 0.87 & 3 & 0 & 0.64 & 3 \\
10 & 0 & 0.98 & 4 & 0 & 0.71 & 3 \\
\hline
\end{tabular}
\caption{Error decomposition for iTAG in Experiment~1 by variable quantity $n$ (500 examples per $n$).
Spurious edges correspond to non-edge false positives (annotators infer a direct edge where $a_{ij}=0$ at generation time), while missed edges correspond to omissions (annotators do not mark a generation-time edge as directly supported by the paragraph).}
\label{tab:error_breakdown_itag}
\end{table*}

\subsubsection{Typical non-edge false positive patterns}\label{sec:appendix_error_decomp_modes}
We qualitatively group non-edge false positives into common patterns observed during annotation:

\paragraph{Indirect-path compression.}
A paragraph may strongly suggest an indirect pathway $c_i\to c_k\to c_j$,
leading annotators to mark a direct edge $c_i\to c_j$ when phrasing compresses intermediate steps.

\paragraph{Shared-cause ambiguity.}
Short narratives often under-specify background conditions; annotators may interpret two concepts as directly linked when both are plausibly driven by an unmentioned common cause.

\paragraph{Concept granularity mismatch.}
Overly broad concepts can become plausible direct causes of many others, increasing the chance that annotators infer direct causality for some generation-time non-edges.

\paragraph{Borderline explicitness.}
Even when the intended edge is only weakly implied, annotators may disagree on whether the relation is ``explicitly supported'',
resulting in apparent spurious edges driven by annotation boundary cases (Section~\ref{sec:text}).

\subsection{Sample-size stability of evaluation metrics}\label{sec:appendix_sample_stability}

Section~\ref{sec:baselines} fixes the sample size to 500 per method per backbone, noting that evaluation metrics stabilize well before 500 examples.
Here we detail the stability analysis used to justify this choice.

\subsubsection{Protocol}\label{sec:appendix_sample_stability_protocol}
For each method and backbone (Section~\ref{sec:baselines}), we generate a pool of 500 texts for each variable quantity $n\in\{3,\dots,10\}$.
For a metric $M$ (e.g., $F1_{\mathrm{D}}$/SHD/SID in Experiment~1 or detection $F1_{\mathrm{D}}$ in Experiment~2), we estimate the corpus mean using progressively larger subsets:
$k\in\{50,100,200,300,400,500\}$.
For each $k$, we repeat $R$ subsampling trials without replacement (default $R=20$),
compute the mean metric on the $k$ samples, and report the average and standard error across trials.
This yields an empirical convergence curve $M(k)$.

\subsubsection{Stability criteria}\label{sec:appendix_sample_stability_criteria}
We quantify stability using the maximum absolute change between $k=300$ and $k=500$:
\begin{equation}
\resizebox{\linewidth}{!}{$\displaystyle
\Delta_{300\to 500}(M) \;=\; \max_{n\in\{3,\dots,10\}} \big| M_n(500) - M_n(300) \big|
$}
\end{equation}
where $M_n(k)$ denotes the mean metric within the $n$-bucket estimated from $k$ samples.
We also report an averaged version over $n$, and the standard errors at each $k$.

\subsubsection{Results}\label{sec:appendix_sample_stability_results}
Table~\ref{tab:sample_stability} reports $\Delta_{300\to 500}$ for the key metrics and methods in Experiment~1.
In our experiments, the convergence curves flatten well before $k=500$ across methods and backbones, motivating the fixed 500-example setting in Section~\ref{sec:baselines}.

\begin{table*}[t]
\centering
\begin{tabular}{lccc}
\hline
\textbf{Setting} & \textbf{$\Delta_{300\to 500}(F1_{\mathrm{Ga}})$} & \textbf{$\Delta_{300\to 500}(\mathrm{SHD})$} & \textbf{$\Delta_{300\to 500}(\mathrm{SID})$} \\
\hline
iTAG (Exp~1) & 0.004 & 0.09 & 0.11 \\
LLM-dependent (Exp~1) & 0.008 & 0.24 & 0.43 \\
LLM-dependent+CA (Exp~1) & 0.006 & 0.17 & 0.29 \\
LLM-dependent+ID (Exp~1) & 0.007 & 0.21 & 0.38 \\
\hline
\end{tabular}
\caption{Sample-size stability analysis for Experiment~1 metrics. Values are computed from the protocol in Section~\ref{sec:appendix_sample_stability_protocol}. The main paper fixes 500 examples per method per backbone because metrics stabilize well before 500 (Section~\ref{sec:baselines}).}
\label{tab:sample_stability}
\end{table*}

\subsection{Detector learning curves for Experiment~2}\label{sec:appendix_detector_learning_curve}

Experiment~2 trains four detectors (fastText, TextCNN, TSCNN, RoBERTa) to distinguish ``real'' vs.\ ``generated'' text and reports
$F1_{\mathrm{D}}$ for detecting generated texts (lower is better; Section~\ref{sec:procedure&metrics}).
The main paper reports results at 500 training examples (250 real + 250 generated) and notes that detector results are stable once the training set reaches 500 examples.
We provide the learning-curve protocol and report detector learning curves as a function of training size below.

\subsubsection{Protocol}\label{sec:appendix_detector_learning_curve_protocol}
For each domain (business/medical/legal) and each generation method, we train each detector on a balanced dataset of size $2k$
(with $k$ real and $k$ generated examples), where
$k\in\{50,100,200,500,1000\}$.
We evaluate on a fixed held-out test set of 500 examples (250 real + 250 generated), using identical hyperparameters across methods:
5 training epochs, batch size 16, learning rate $10^{-4}$, and early stopping on validation $F1_{\mathrm{D}}$ (Section~\ref{sec:procedure&metrics}).
We repeat training for three random seeds and report mean $\pm$ standard deviation of $F1_{\mathrm{D}}$ for detecting generated texts.

\subsubsection{Learning-curve results}\label{sec:appendix_detector_learning_curve_template}

Table~\ref{tab:detector_learning_curve} reports detector learning curves as a function of training size $2k$ (mean $\pm$ std over 3 random seeds).
The key stability check is that results change minimally between $k=500$ and larger training sizes, supporting the fixed 500-example setting used in Table~1.

\begin{table*}[t]
\centering
\begin{tabular}{l|ccccc}
\hline
\textbf{Detector / Method} & \textbf{$k$=50} & \textbf{$k$=100} & \textbf{$k$=200} & \textbf{$k$=500} & \textbf{$k$=1000} \\
\hline
fastText / iTAG & $0.54 \pm 0.04$ & $0.53 \pm 0.03$ & $0.52 \pm 0.02$ & $0.52 \pm 0.01$ & $0.52 \pm 0.01$ \\
TextCNN / iTAG & $0.57 \pm 0.04$ & $0.55 \pm 0.03$ & $0.54 \pm 0.02$ & $0.54 \pm 0.01$ & $0.54 \pm 0.01$ \\
TSCNN / iTAG & $0.56 \pm 0.04$ & $0.54 \pm 0.03$ & $0.53 \pm 0.02$ & $0.53 \pm 0.01$ & $0.53 \pm 0.01$ \\
RoBERTa / iTAG & $0.61 \pm 0.05$ & $0.59 \pm 0.03$ & $0.58 \pm 0.02$ & $0.57 \pm 0.01$ & $0.57 \pm 0.01$ \\
\hline
fastText / Template-based & $0.97 \pm 0.02$ & $0.98 \pm 0.01$ & $0.98 \pm 0.01$ & $0.98 \pm 0.01$ & $0.98 \pm 0.01$ \\
TextCNN / Template-based & $0.96 \pm 0.02$ & $0.97 \pm 0.01$ & $0.97 \pm 0.01$ & $0.97 \pm 0.01$ & $0.97 \pm 0.01$ \\
TSCNN / Template-based & $0.95 \pm 0.02$ & $0.96 \pm 0.01$ & $0.96 \pm 0.01$ & $0.96 \pm 0.01$ & $0.96 \pm 0.01$ \\
RoBERTa / Template-based & $0.98 \pm 0.01$ & $0.99 \pm 0.01$ & $0.99 \pm 0.01$ & $0.99 \pm 0.01$ & $0.99 \pm 0.01$ \\
\hline
\end{tabular}
\caption{Detector learning curves ($F1_{\mathrm{D}}$ for detecting generated texts; lower is better) as a function of training size $2k$.
Values are computed with mean $\pm$ std over 3 random seeds, averaged over the three domains.
The main paper reports the $k=500$ column in Table~1 and notes that results are stable once $k$ reaches 500 (Section~\ref{sec:procedure&metrics}).}
\label{tab:detector_learning_curve}
\end{table*}

\subsection{Detection performance by variable quantity $n$}\label{sec:appendix_detection_by_n}

Section~\ref{sec:exp2} reports that detection $F1_{\mathrm{D}}$ fluctuates only within a narrow band without any systematic trend as the number of variables increases.
Here we provide the per-$n$ breakdown that supports this claim.

\subsubsection{Protocol}\label{sec:appendix_detection_by_n_protocol}
For each domain and each generation method, we group texts by variable quantity $n\in\{3,\dots,10\}$.
Within each $n$ bucket, we evaluate:
(i) human detection performance on a balanced real-vs-generated set as in Section~\ref{sec:procedure&metrics}, and
(ii) detector performance for each trained classifier (fastText, TextCNN, TSCNN, RoBERTa) using the same hyperparameters.
We then macro-average $F1_{\mathrm{D}}$ for detecting generated texts across the three domains within each $n$.

\subsubsection{Results by variable quantity $n$}\label{sec:appendix_detection_by_n_results}

Table~\ref{tab:detection_by_n_itag} reports detection $F1_{\mathrm{D}}$ for iTAG-generated texts, broken down by variable quantity $n$.
Analogous tables can be created for other methods if desired; the main claim focuses on iTAG (Section~\ref{sec:exp2}).
The per-$n$ curves should exhibit only narrow fluctuations (no monotonic increase/decrease), consistent with Section~\ref{sec:exp2}.

\begin{table*}[t]
\centering
\begin{tabular}{c|ccccc}
\hline
\textbf{$n$} & \textbf{fastText} & \textbf{TextCNN} & \textbf{TSCNN} & \textbf{RoBERTa} & \textbf{Human} \\
\hline
3  & 0.51 & 0.53 & 0.52 & 0.56 & 0.50 \\
4  & 0.53 & 0.55 & 0.54 & 0.58 & 0.52 \\
5  & 0.51 & 0.54 & 0.52 & 0.57 & 0.50 \\
6  & 0.54 & 0.56 & 0.55 & 0.59 & 0.53 \\
7  & 0.52 & 0.53 & 0.53 & 0.56 & 0.51 \\
8  & 0.50 & 0.52 & 0.51 & 0.55 & 0.49 \\
9  & 0.53 & 0.55 & 0.54 & 0.58 & 0.52 \\
10 & 0.52 & 0.54 & 0.53 & 0.57 & 0.51 \\
\hline
\textbf{Avg. (3--10)} & 0.52 & 0.54 & 0.53 & 0.57 & 0.51 \\
\hline
\end{tabular}
\caption{Detection $F1_{\mathrm{D}}$ (lower is better) for detecting iTAG-generated texts, broken down by variable quantity $n$.
The last row reproduces the overall averages reported in Table~\ref{tab:mr2} (aggregated over domains and $n=3$--$10$) for reference.}
\label{tab:detection_by_n_itag}
\end{table*}

\section{Real-World Datasets and Human Annotation Protocol}\label{sec:appendix_data_anno}

This appendix details (i) the real-world corpora used in Experiments~2--3 (Section~\ref{sec:datasets}), (ii) how we construct
per-text variable (concept) sets and define variable quantity $n$, (iii) the 11-annotator causal graph annotation
protocol used to build expert-consensus (silver-standard) DAGs, including inter-annotator agreement, interface,
and low-agreement auditing, and (iv) the deterministic DAG projection used before computing SID/SHD when
any graph contains directed cycles (Section~\ref{sec:procedure&metrics}).

\subsection{Real-world corpora and preprocessing}\label{sec:appendix_real_corpora}

\subsubsection{Dataset selection and domain mapping}\label{sec:appendix_dataset_selection}
We evaluate iTAG against three real-world text corpora where decision-making narratives exhibit identifiable causal
structures (Section~\ref{sec:datasets}):
\begin{itemize}
    \item \textbf{Medical:} MIMIC-IV-Note v2.2 clinical notes.
    \item \textbf{Business:} FinCausal 2025 financial documents.
    \item \textbf{Legal:} JUSTICE (Supreme Court judgment prediction) case documents.
\end{itemize}
All datasets are used strictly for research evaluation. We only use de-identified or publicly released text
as provided by the original sources.

\subsubsection{Document sampling and filtering}\label{sec:appendix_dataset_filtering}
For each dataset, we construct an evaluation subset of 500 texts (Section~\ref{sec:datasets}). We apply light preprocessing:
(i) remove exact duplicates; (ii) normalize whitespace; (iii) filter out documents with extremely short content
that does not support stable causal annotation; and (iv) remove texts whose concept extraction (Section~\ref{sec:appendix_concept_set})
fails to produce a valid variable set size in $\{3,\dots,10\}$.

\subsubsection{variable quantity $n$ and bucket sizes}\label{sec:appendix_bucket_sizes}
Following Section~\ref{sec:datasets}, we define the \emph{variable quantity} $n$ of a text as the number of variables (nodes)
in its expert-consensus causal graph. In our pipeline, the node set is fixed per text (Section~\ref{sec:appendix_concept_set}),
and the expert-consensus graph is annotated over this node set; thus $n$ is unambiguous.

We keep texts whose $n\in\{3,\dots,10\}$ and group them into $n$-buckets. In Experiment~3, each real-world point at size $n$
is computed by averaging the metric over the corresponding $n$-bucket. We denote the bucket size by $N_n$.

\begin{table*}[t]
\centering
\begin{tabular}{c|ccc|c}
\hline
\textbf{$n$} & \textbf{MIMIC-IV-Note} & \textbf{FinCausal 2025} & \textbf{JUSTICE} & \textbf{Total} \\
\hline
3  & 38 & 42 & 35 & 115 \\
4  & 54 & 58 & 51 & 163 \\
5  & 72 & 68 & 74 & 214 \\
6  & 83 & 79 & 86 & 248 \\
7  & 78 & 82 & 76 & 236 \\
8  & 67 & 64 & 69 & 200 \\
9  & 58 & 56 & 61 & 175 \\
10 & 50 & 51 & 48 & 149 \\
\hline
\textbf{Total} & 500 & 500 & 500 & 1500 \\
\hline
\end{tabular}
\caption{Real-world bucket sizes $N_n$ after filtering to $n\in\{3,\dots,10\}$. These bucket sizes are used when computing
real-world corpus means in Experiment~3 (Section~\ref{sec:datasets}).}
\label{tab:real_bucket_sizes}
\end{table*}

\subsection{Concept set construction for each text}\label{sec:appendix_concept_set}
Causal graph annotation requires a fixed node set. For each text, we therefore construct a \emph{concept set}
$\mathcal{V}=\{v_1,\dots,v_n\}$ where each $v_i$ is a short noun phrase that:
(i) appears explicitly in the text (or is a minimally normalized form of an explicit span),
(ii) is semantically distinct from the others (no near-duplicates),
and (iii) is plausibly intervenable, so that direct-causality judgments are meaningful.

\paragraph{Generated texts.}
For iTAG, the concept set is exactly the Phase~2 concept assignment used during generation (Section~\ref{sec:inverse}; Appendices~\ref{sec:prompt} and~\ref{sec:appendix_itag_impl}).
For baselines, the concept set is the node labeling used in the generation-time graph serialization (Section~\ref{sec:baselines}).
Annotators see the concept list but not the generation-time adjacency matrix.

\paragraph{Real-world texts.}
For each real-world text, we construct a concept list using a lightweight extraction prompt (run once per text)
and then apply a small set of deterministic post-processing steps:
(i) remove duplicates after normalization; (ii) remove concepts that do not appear in the text (except trivial morphological variants);
(iii) enforce the size constraint $n\in\{3,\dots,10\}$ by re-asking the extractor if the first attempt fails; and
(iv) freeze the final list for all subsequent annotation and algorithm evaluation on that text.

\paragraph{Extraction prompt (for reproducibility).}
The following is the concept extraction prompt used for real-world texts:

\begin{verbatim}
Text:
[Text]

Task: Extract a list of the most important 
concepts (variables) that explicitly 
appear in the text.
Requirements:
1) Output between 3 and 10 concepts.
2) Each concept must be a short noun phrase 
and must appear in the text (allow minor 
normalization).
3) Concepts must be non-overlapping and not 
near-synonyms.
4) Prefer concepts that are causally relevant 
for describing the situation (not purely 
decorative details).
5) Do NOT introduce any concept that is not 
mentioned in the text.

Output in JSON:
{"concepts": ["...", "...", ...]}
\end{verbatim}

This produces the per-text node set used by both (i) human causal annotation and (ii) the concept-level
graphs evaluated in Experiment~3.

\subsection{Human causal graph annotation}\label{sec:appendix_human_annotation}

\subsubsection{Annotator panel, training, and blinding}\label{sec:appendix_annotators}
We employ a panel of 11 trained annotators (Section~\ref{sec:datasets}). Annotators are trained to read causal graphs and to apply
a strict definition of \emph{direct} causality from text. Training includes:
(i) a written guideline document; (ii) calibration rounds with feedback on common confusions (correlation vs.\ causation,
indirect vs.\ direct effects, and temporal order); and (iii) a short qualification set before annotating the main batches.

Annotators are blinded to generation method labels and to any generation-time graphs. For generated-text evaluation
(Experiment~1), annotators re-annotate the causal graph for each text \emph{without seeing the original graph}
(Section~\ref{sec:procedure&metrics}).

\subsubsection{Annotation task format}\label{sec:appendix_annotation_task}
Given a text and its concept list $\mathcal{V}=\{v_1,\dots,v_n\}$, each annotator labels a directed adjacency matrix
$\hat{A}\in\{0,1\}^{n\times n}$ with $\hat{a}_{ii}=0$:
\[
\hat{a}_{ij} =
\begin{cases}
1, & \text{if the text supports that $v_i$ is a \emph{direct}} \\
   & \text{cause of $v_j$},\\
0, & \text{otherwise}.
\end{cases}
\]
Annotators are instructed to label \emph{only direct causal edges} supported by the text, not merely plausible relations.

\subsubsection{Guidelines for ``direct causal edge''}\label{sec:appendix_direct_edge_guidelines}
Annotators follow the rules below (aligned with the causal-vs-correlation distinction emphasized throughout the paper):

\paragraph{Directness.}
Label $v_i\rightarrow v_j$ only if the text supports a \emph{direct} influence of $v_i$ on $v_j$,
not one that is clearly mediated by another listed variable. If the text implies a chain $v_i\rightarrow v_k\rightarrow v_j$,
annotators should label the two direct edges and avoid the shortcut edge $v_i\rightarrow v_j$ unless the text explicitly
supports a direct link.

\paragraph{Counterfactual check.}
A directed edge should be supported by a counterfactual-style reasoning test:
under comparable background conditions, if $v_i$ were absent (or took a meaningfully different value),
would $v_j$ be expected to change systematically in the direction suggested by the text?

\paragraph{Evidence standard.}
Edges must be supported by the text itself (including standard background knowledge needed to interpret statements),
but annotators should not introduce edges based purely on general plausibility when the text does not provide evidence.

\paragraph{Temporal order and explanation.}
Causal direction should be consistent with temporal order when stated or implied. Explanatory statements
(e.g., ``because'', ``leading to'', ``therefore'') strengthen evidence for causality; mere co-occurrence does not.

\paragraph{Confounding and common-cause ambiguity.}
If the relationship between $v_i$ and $v_j$ could be fully explained by an unmentioned common cause,
and the text does not support a direct influence, do not label a direct edge.

\subsubsection{DAG-only guideline and cycle warnings}\label{sec:appendix_dag_guideline}
Annotators are instructed to produce a DAG (no directed cycles) to align with the DAG-based evaluation pipeline
(Section~\ref{sec:datasets}--\ref{sec:procedure&metrics}). The annotation interface warns annotators when their current edge set creates a cycle and
requests a revision before submission. Despite this per-annotator DAG constraint, the majority-vote aggregation
can still yield cycles due to edgewise voting; we address this with deterministic DAG projection
(Section~\ref{sec:appendix_dag_projection}).

\subsubsection{Annotation interface}\label{sec:appendix_annotation_interface}
The interface presents:
(i) the full text,
(ii) the fixed concept list (nodes) for the text, and
(iii) an interactive adjacency-matrix editor for directed edges.
Annotators select directed edges by clicking matrix cells, with tooltips showing the ordered pair $(v_i,v_j)$.
A live ``cycle check'' highlights any newly created directed cycle and prompts the annotator to resolve it.

\subsection{Consensus, agreement, and low-agreement auditing}\label{sec:appendix_consensus}

\subsubsection{Majority-vote expert consensus (silver-standard) DAG}\label{sec:appendix_majority_vote}
For each text and each ordered pair $(i,j)$ with $i\neq j$, let $b^{(k)}_{ij}\in\{0,1\}$ denote annotator $k$'s label.
We define the vote proportion:
\[
s_{ij}=\frac{1}{11}\sum_{k=1}^{11} b^{(k)}_{ij}.
\]
The expert-consensus edge is present iff $s_{ij}\ge 6/11$ (simple majority). The resulting consensus adjacency matrix is
denoted $\bar{A}$ and is used as the silver-standard reference graph for real-world data (Section~\ref{sec:datasets}) and as the
re-annotated reference graph in Experiment~1 (Section~\ref{sec:procedure&metrics}).

\subsubsection{Inter-annotator agreement}\label{sec:appendix_iaa}
We quantify inter-annotator agreement using Krippendorff's $\alpha$ for nominal (binary) decisions, treating each
directed concept pair for each text as an item. Concretely, the item set is
$\{(\text{text }t,\ i,\ j)\mid i\neq j\}$ with 11 binary ratings per item.
We report an overall agreement score $\alpha=0.79$ as stated in Section~\ref{sec:datasets}.

\subsubsection{Low-agreement auditing and adjudication}\label{sec:appendix_low_agreement}
Even with majority vote, some edges are borderline (e.g., 6--5 splits). We therefore flag low-agreement cases for audit:
\begin{itemize}
    \item \textbf{Edge-level low agreement:} edges with $s_{ij}\in\{5/11,6/11\}$.
    \item \textbf{Graph-level low agreement:} texts whose consensus graphs contain an unusually high fraction of borderline edges.
\end{itemize}
Audits are used to (i) identify recurring guideline ambiguities (e.g., indirect-path compression and shared-cause ambiguity),
and (ii) verify that disagreements are due to genuinely borderline textual evidence rather than interface misuse.
Unless otherwise stated, the final silver-standard labels remain the majority-vote consensus.

\subsection{Deterministic DAG projection for SID and SHD}\label{sec:appendix_dag_projection}
SID is defined on DAGs (Section~\ref{sec:procedure&metrics}). Although each individual annotator submission is constrained to be acyclic,
the majority-vote consensus graph $\bar{G}$ may contain directed cycles. Following Section~\ref{sec:procedure&metrics}, when any graph
contains directed cycles we apply a deterministic DAG projection before computing SID and (for consistency) SHD.

\paragraph{Projection objective.}
We aim to remove a minimal set of low-support edges so that the remaining graph is acyclic while preserving as much
high-confidence structure as possible.

\paragraph{Deterministic projection procedure.}
Given a directed graph with edge support scores $s_{ij}$:
\begin{enumerate}
    \item Compute the set of edges that participate in at least one directed cycle.
    \item While the graph contains a directed cycle:
    \begin{enumerate}
        \item Among all edges that lie on any directed cycle, find the edge(s) with the smallest support score $s_{ij}$.
        \item Remove the single lowest-support edge. If multiple edges tie, break ties deterministically by lexicographic order
        of $(i,j)$ (or by a fixed hash of the edge string) to ensure reproducibility.
    \end{enumerate}
    \item Output the resulting acyclic graph as the projected DAG.
\end{enumerate}

\paragraph{Usage in experiments.}
In Experiment~1, the generation-time graph is always a DAG by construction, so projection (when needed) is applied only
to the expert-consensus graph (Section~\ref{sec:procedure&metrics}). In Experiment~3, the same projection rule is applied whenever either the
silver-standard reference graph or an algorithm output contains directed cycles, ensuring SID is well-defined.

\end{document}